\documentclass[10pt,journal,compsoc]{IEEEtran}
\usepackage[nocompress]{cite}
\usepackage{tabu}
\usepackage[margin=0pt,font=small,labelfont=bf,labelsep=endash,tableposition=top]{caption}

\usepackage{amsmath}
\usepackage{array}
\usepackage{booktabs}
\usepackage{fixltx2e}
\usepackage[table]{xcolor}
\usepackage{float}
\usepackage{url}
\usepackage{graphicx}
\usepackage{amsmath}
\usepackage{amssymb}
\usepackage{epsfig}
\usepackage{epstopdf}
\usepackage{makecell,multirow,diagbox}
\usepackage{color}
\usepackage{soul}
\usepackage{url}
\usepackage{amsmath}
\usepackage{array}
\usepackage[utf8]{inputenc}
\usepackage{xcolor}
\usepackage{hyperref}
\usepackage{helvet}
\usepackage{courier}
\usepackage{bm}
\usepackage{bbm}
\usepackage{wrapfig}
\usepackage{picinpar}
\usepackage{arydshln}
\usepackage{algorithm}
\usepackage{algorithmic}
\usepackage{cite}
\usepackage{balance}
\usepackage{amssymb}
\usepackage{multicol}
\usepackage{multirow}
\usepackage{makecell}
\usepackage{mathtools,xspace}
\usepackage{ragged2e}
\usepackage{tabu}
\usepackage{booktabs, multirow, hhline}
\definecolor{grayline}{RGB}{128,128,128}

\hypersetup{
    colorlinks=true,
    breaklinks=true,
    urlcolor= blue,
    linkcolor= red,
    bookmarksopen=false,
    filecolor=black,
    citecolor=blue,
    linkbordercolor=blue
}
\AtBeginDocument{\hypersetup{pdfborder={0 0 1}}}%

\usepackage{color}

\newlength\savewidth
  
\usepackage{xspace}
\makeatletter
\newcommand{\thickhline}{%
    \noalign {\ifnum 0=`}\fi \hrule height 0.7pt
    \futurelet \reserved@a \@xhline
}

\newcolumntype{I}{!{\vrule width 0.8pt}}

\usepackage{graphicx}
\usepackage{float}
\usepackage{subfig}
\usepackage{pifont}
\definecolor{grey}{RGB}{128,128,128}

\newcolumntype{C}[1]{>{\PreserveBackslash\centering}p{#1}}
\usepackage{array}
\newcommand{\PreserveBackslash}[1]{\let\temp=\\#1\let\\=\temp}
\newcolumntype{C}[1]{>{\PreserveBackslash\centering}p{#1}}
\newcolumntype{R}[1]{>{\PreserveBackslash\raggedleft}p{#1}}
\newcolumntype{L}[1]{>{\PreserveBackslash\raggedright}p{#1}}

\definecolor{lightblue}{rgb}{0.21,0.49,0.74}

\begin{document}
\title{Achieving Text-based Person Retrieval \\with Any Granularity}

\author{
Jialong Zuo,
Hanyu Zhou,
Dongyue Wu,
Yongtai Deng,
Mengdan Tan,
Nong Sang,~\IEEEmembership{Member,~IEEE},\\
Changxin Gao,~\IEEEmembership{Senior Member,~IEEE},
Xiang Bai,~\IEEEmembership{Fellow,~IEEE}

\thanks{This work was supported by the National Natural Science Foundation of China No.62176097 and Hubei Provincial Natural Science Foundation of China No.2022CFA055.
(\textit{Corresponding Author: Changxin Gao})}
\thanks{
J.~Zuo, H.~Zhou, D.~Wu, Y.~Deng, M.~Tan, N.~Sang and C.~Gao are with the National Key Laboratory of Multispectral Information Intelligent Processing Technology, School of Artificial Intelligence and Automation, Huazhong University of Science and Technology, Wuhan 430074, China (e-mail: {jlongzuo, nsang, cgao}@hust.edu.cn).
}
\thanks{X.~Bai is with the School of Software Engineering, Huazhong University of Science and Technology, Wuhan 430074, China. (e-mail: xbai@hust.edu.cn).}

\thanks{Open Source Repository: https://github.com/zplusdragon/CMAM}
}

\IEEEtitleabstractindextext{%
\begin{abstract}
\justifying
Text-based person retrieval faces a critical but under-explored challenge: the inherent uncertainty of query granularity in real-world scenarios. This paper introduces a new paradigm, Text-based Person Retrieval with Any Granularity, and provides a systematic solution. First, we formalize a five-level granularity spectrum and construct UFine6926-MG, a high-quality multi-grained dataset annotated comprehensively at all granularities via a novel Multi-grained Text Annotation Engine. Second, acknowledging that coarse queries naturally correspond to multiple valid candidates, we propose MG-Eval, a holistic evaluation benchmark with progressively detailed texts and cross-identity labels that reflect real-world semantics, alongside tailored evaluation metrics and protocols. Third, after a comprehensive diagnosis reveals the systemic limitations of existing research, we propose the Cross-modal Multi-grained Aligning and Matching (CMAM) framework. CMAM achieves granularity-aware retrieval through: 1) orthogonal-expert perception to disentangle granularity-specific features; 2) probabilistic alignment to model many-to-many matches under query uncertainty; and 3) granularity-consistent reasoning to steer feature learning via joint cross-modal granularity verification. Experiments demonstrate that CMAM significantly outperforms state-of-the-art methods across all granularity levels. This work establishes a foundational benchmark and a robust baseline, paving the way for more practical person retrieval systems.

\end{abstract}

}

\maketitle

\IEEEdisplaynontitleabstractindextext
\IEEEpeerreviewmaketitle

\IEEEraisesectionheading{\section{Introduction}\label{sec:introduction}}

\IEEEPARstart{T}{ext-based} person retrieval (TPR)~\cite{cuhkpedes} aims to search for a target person from a large-scale image gallery based on a free-form textual description. Unlike traditional person re-identification that relies on visual matching~\cite{zheng2015scalable,wei2018person,he2021transreid,zuo2024cross}, this task requires a deep semantic understanding of both the image content and the textual query, and more importantly, the fine-grained alignment between the two modalities. This technology has wide-ranging practical applications, from video surveillance and public security to social media analysis, offering a flexible and intuitive interface for human-computer interaction~\cite{ye2021deep,ye2025transformer}.

Despite significant progress, existing research has predominantly been conducted under a constrained and unrealistic setting: the assumption of fixed textual granularity. Most benchmarks and methods are designed for and evaluated on either coarse-grained attribute-like queries~\cite{cuhkpedes,icfgpedes,rstpreid} or, more recently, ultra-fine-grained detailed descriptions~\cite{ufinebench}. This narrow focus creates a fundamental gap between academic research and the real world, where the granularity of a textual query is inherently uncertain, diverse, and unpredictable. The critical challenge of \textit{any granularity retrieval} has thus been largely overlooked.

This oversight leads to two critical and interconnected issues that severely hinder practical deployment. First, from the model capability perspective, the real world presents a complex spectrum of query specificity. A user might begin with a vague search and progressively refine it based on initial results~\cite{bai2025chat,lu2025llavareid,qin2025human}. Models trained on fixed-granularity data lack this adaptability; they are brittle when faced with granularity shifts, leading to a sharp decline in performance outside their training domain. Second, from the evaluation perspective, the prevailing assessment paradigm is fundamentally misaligned with real-world semantics. The strict one-text-to-one-identity binding~\cite{cuhkpedes} fails to acknowledge that a coarse-grained description can legitimately correspond to multiple visually similar individuals. By penalizing models for retrieving these semantically valid matches, current protocols provide an inaccurate measure of a model's true retrieval capability, especially in multi-grained scenarios. Therefore, a comprehensive exploration into \textit{Text-based Person Retrieval with Any Granularity} is not merely an academic refinement but a necessary step toward building robust and applicable person retrieval systems.

To bridge this gap, as shown in Fig.~\ref{fig:main}, this work establishes a foundational benchmark and provides a systematic solution through the following three core contributions:

\begin{figure*}[htb]
\centering
\includegraphics[width=\linewidth]{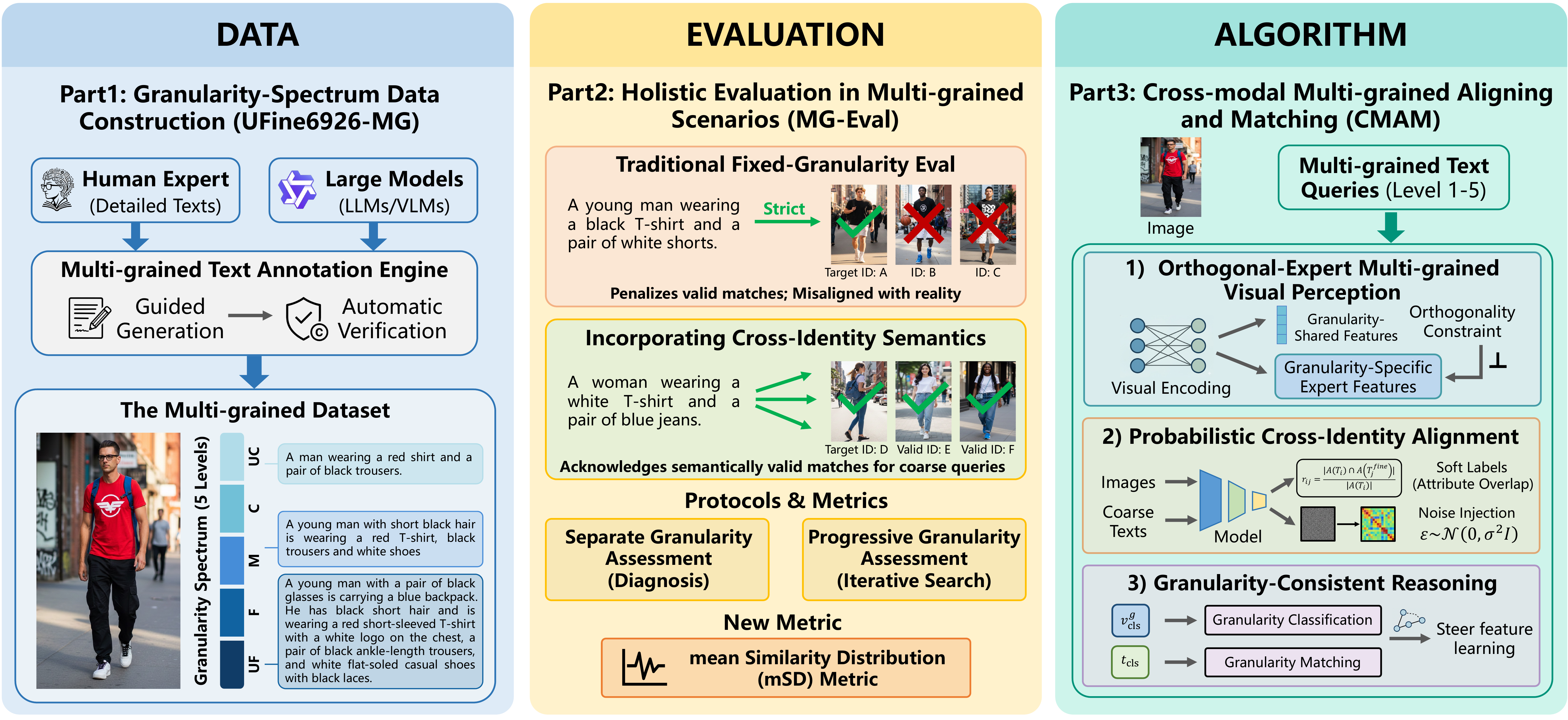}
\vspace{-7mm}
\caption{An overview of the proposed systematic paradigm for Text-based Person Retrieval with Any Granularity. Our contribution constitutes a holistic solution: 1) Data: The construction of the UFine6926-MG dataset via a Multi-grained Text Annotation Engine, establishing a comprehensive granularity spectrum. 2) Evaluation: The MG-Eval benchmark incorporating cross-identity semantics and progressive-granularity protocols to reflect real-world retrieval complexity. 3) Algorithm: The CMAM framework, which features orthogonal-expert perception, probabilistic cross-identity alignment, and granularity-consistent reasoning.}
\vspace{-4mm}
\label{fig:main}
\end{figure*}

\textbf{Data}. We establish a rigorous granularity spectrum and construct a high-quality multi-grained dataset to address the data bottleneck. Moving beyond simplistic text length metrics, we define five granularity levels based on the number of semantic attribute types, providing a principled measure of textual informativeness. To enable efficient large-scale annotation, we design a Multi-grained Text Annotation Engine that synergizes human expertise with large models~\cite{bai2025qwen2,yang2025qwen3}. Starting from manually annotated detailed texts~\cite{ufinebench}, this engine employs guided LLMs and VLMs to automatically generate and verify descriptions across all granularity levels. The result is UFine6926-MG, a comprehensive dataset where each image is annotated with granularity-controlled texts, serving as an essential resource for training and diagnosing models capable of handling granularity diversity.

\textbf{Evaluation}. We propose MG-Eval, a holistic multi-grained evaluation benchmark designed to reflect real-world retrieval semantics by incorporating cross-identity labels, acknowledging that coarse-grained queries naturally correspond to multiple valid candidates. This benchmark features two tailored evaluation protocols: separate granularity assessment for granularity-wise diagnosis, and progressive granularity assessment simulating iterative search process~\cite{lu2025llavareid}. As an integral component, we further introduce the mean Similarity Distribution (mSD) metric. Specifically for multi-grained scenarios, mSD leverages continuous similarity distributions to provide a more sensitive and precise measurement of a model's ability to discern fine-grained differences across varying levels of textual granularity.

\textbf{Algorithm}. Through a comprehensive diagnosis, we reveal the systemic limitations of existing research~\cite{cuhkpedes,icfgpedes,irra,rde,xie2025fgclip} in handling granularity variation. In response, we propose a Cross-modal Multi-grained Aligning and Matching (CMAM) framework, a comprehensive architecture designed to explicitly model and adapt to the granularity spectrum. The framework is built upon three core innovations: 1) An Orthogonal-Expert Multi-grained Visual Perception module employs structurally heterogeneous experts to extract discriminative features across granularities. It integrates a shared transformation capturing common patterns with granularity-specific pathways, regulated by an orthogonality loss to ensure feature diversity and disentanglement. 2) A Probabilistic Cross-Identity Alignment mechanism addresses the inherent many-to-many matching of coarse-grained texts. We construct soft labels based on attribute overlap between coarse queries and fine-grained texts of other identities, with Gaussian noise simulating real-world ambiguity to enable robust probabilistic training. 3) A Granularity-Consistent Reasoning module, which performs reasoning to assess image-text granularity consistency, provides fine-grained signals to steer the granularity-aware feature learning process.

Extensive experiments validate the superiority of our framework. While achieving competitive results on conventional fixed-granularity benchmarks~\cite{cuhkpedes,icfgpedes,ufinebench}, CMAM exhibits a more pronounced performance lead in multi-grained evaluation settings. Furthermore, CMAM demonstrates exceptional training stability when dealing with diverse granularity distributions, a critical advantage where traditional methods often suffer from performance decline. Qualitative analyses further confirm that CMAM effectively extracts discriminative visual cues tailored to varying granularities, underscoring its superior representation robustness and practical utility in complex real-world scenarios.

Overall, the main contributions of this paper include:

(1) The formalization of \textit{any granularity retrieval} as a new paradigm for text-based person retrieval, addressing the critical gap between constrained benchmarks and real-world scenarios with varying query granularities.

(2) The construction of UFine6926-MG, the first multi-grained dataset with granularity-controlled annotations, and the proposal of MG-Eval, a holistic evaluation benchmark with progressively detailed texts and cross-identity labels, alongside tailored evaluation metrics and protocols.

(3) The proposal of the CMAM framework, which introduces three novel technical innovations: an Orthogonal-Expert module for granularity-disentangled feature learning, a Probabilistic Alignment mechanism for cross-identity matching, and a Granularity-Consistent Reasoning module for joint cross-modal granularity verification.

(4) Extensive experiments demonstrating state-of-the-art performance across both existing benchmarks and our proposed multi-grained evaluation suite, with both quantitative and qualitative evidence confirming CMAM's granularity-aware representation learning capability.

This work is an extension of our CVPR conference paper~\cite{ufinebench}. It advances the research topic from a singular focus on ultra fine-grained retrieval to the more challenging and realistic problem of \textit{any granularity retrieval}. The key improvements include: 1) We newly define a granularity spectrum and develop a Multi-grained Text Annotation Engine, evolving the original UFine6926 into UFine6926-MG, a multi-grained dataset with granularity-controlled annotations. 2) We introduce MG-Eval, a holistic evaluation benchmark specifically designed for assessing multi-grained retrieval capability. It deeply considers the inherent ambiguity of coarse-grained texts by pioneering the incorporation of cross-identity labels, and better aligns with real-world semantics through two novel protocols. 3) We propose the CMAM framework, which is fundamentally redesigned with three novel components: an Orthogonal-Expert module for granularity-disentangled feature learning, a Probabilistic Alignment mechanism for cross-identity matching, and a Granularity-Consistent Reasoning module. These enhancements explicitly address granularity variation and substantially outperform our prior CFAM model. 4) We expand extensive experiments that systematically diagnose existing research in terms of both data and methods, providing comprehensive evidence of the effectiveness and generalizability of our solutions under diverse settings.

\vspace{-2mm}
\section{Related Work}
\textbf{Text-based Person Retrieval.} Dataset evolution in TBPR reflects a growing demand for fine-grained discriminative power in real-world scenarios. CUHK-PEDES~\cite{cuhkpedes} established the field foundation by introducing natural language descriptions. ICFG-PEDES~\cite{icfgpedes} further emphasized identity-centric features with longer descriptions and complex backgrounds. RSTPReID~\cite{rstpreid} focused on decoupling person-specific information from environmental noise across multi-camera views. More recently, UFine6926~\cite{ufinebench} pushed toward ultra-fine granularity with exceptionally detailed descriptions averaging 80.8 words. Additionally, synthetic datasets such as SYNTH-PEDES~\cite{zuo2024plip} and MALS~\cite{mals} leverage generative models to support large-scale pre-training.

Meanwhile, algorithmic progress has transitioned from global feature matching~\cite{zhang2018deep,zheng2020dual,chen2018improving,wang2019language} to fine-grained local alignment~\cite{jing2020pose,wang2020vitaa,niu2020improving,wang2022caibc,wu2021lapscore} and foundation model adaptation~\cite{irra,rde,cao2024empirical,yan2023clip,wang2026p}. Early approaches employed gated recurrent attention for holistic image-text association~\cite{cuhkpedes}, followed by explicit structural alignment via hierarchical part-based partitioning~\cite{niu2020improving} or feature decomposition to disentangle identity-specific cues from complex contexts~\cite{icfgpedes, yu2026i2id}. The advent of CLIP~\cite{clip} marked a paradigm shift toward implicit reasoning, with IRRA~\cite{irra} establishing fine-grained region-word correspondences without auxiliary detectors, and subsequent works exploring fine-grained semantics-aware representation learning~\cite{wang2024fine}, parameter-efficient fine-tuning~\cite{liu2024clip}, and visual feature enhancement~\cite{shen2025enhancing}. More recently, conversational retrieval with multi-turn dialogues has further extended the task scope~\cite{bai2025chat,qin2025human,lu2025llavareid}. Nevertheless, even methods incorporating global-local alignment remain constrained by the fixed-granularity assumption, lacking systematic exploration of variable-granularity inputs, which is a critical bottleneck that underscores the necessity for adaptive granularity mechanisms.

\noindent
\textbf{Granularity Modeling in Image-Text Retrieval.} Granularity modeling has emerged as a critical dimension in advancing image-text retrieval. Early approaches~\cite{faghri2018vse++,clip,li2021align,li2022blip} predominantly adopted coarse-grained paradigms, representing entire images and sentences as single global embeddings within a shared semantic space. Representative works such as VSE++~\cite{faghri2018vse++} introduced hard negative mining with triplet ranking losses to enhance discriminative capability, while CLIP~\cite{clip} leveraged large-scale contrastive pre-training on web-scale image-text pairs to achieve remarkable zero-shot transferability. While efficient and scalable, these global alignment methods neglect local semantic structures, often failing to capture object relationships and attribute distinctions in complex scenes.

To address these limitations, subsequent research~\cite{yao2022filip,zhang2024long,chen2020imram,pan2023fine,xie2025fgclip,xiao2025flair} shifted toward fine-grained alignment strategies that explicitly model correspondences between image regions and textual words or phrases. Methods like IMRAM~\cite{chen2020imram} introduced iterative matching mechanisms with recurrent attention memory to progressively refine region-word alignments, while CHAN~\cite{pan2023fine} proposed hard assignment encoding to replace soft attention, reducing computational overhead while preserving local matching signals. More recently, FGCLIP~\cite{xie2025fgclip} and FLAIR~\cite{xiao2025flair} further advance this direction by incorporating large-scale fine-grained annotations and long-form texts, enabling precise localization of detailed concepts through text-conditioned attention pooling. However, these approaches often incur substantial computational costs and remain susceptible to noise from erroneous region proposals or misalignments.

The current frontier focuses on multi-granularity modeling~\cite{hendriksen2025benchmark}, seeking to unify coarse and fine-grained representations for balanced efficiency and precision. X-VLM~\cite{zeng2022multi} pioneered this by jointly learning image, region, and patch-level concepts. Subsequent approaches like PixCLIP~\cite{xiao2025pixclip} and MulCLIP~\cite{truong2025mulclip} enable dynamic adaptation to varying query complexities without external detectors, while task-specific extensions (e.g., X-CLIP~\cite{ma2022x} and MGSGM~\cite{huang2025mgsgm}) demonstrate the efficacy of cross-grained contrastive learning. Despite this progress, developing adaptive granularity selection, uncertainty-aware alignment for robustness, and comprehensive multi-grained evaluation benchmarks in TBPR remain significant challenges.

It is worth noting that our work differs fundamentally from existing multi-granularity works such as X-VLM~\cite{zeng2022multi} and PixCLIP~\cite{xiao2025pixclip} in three aspects: (1) we define granularity over \textit{textual semantic density} of person-centric attributes rather than visual spatial hierarchies; (2) we target \textit{granularity-adaptive retrieval under query uncertainty} rather than general multi-level representation learning; and (3) we explicitly address the \textit{one-to-many correspondence} inherent in coarse-grained person retrieval, where a vague query legitimately matches multiple identities. These distinctions motivate a technically unique solution at the data, evaluation, and algorithmic levels.

\newpage
\newpage

\begin{figure}[htb]
\centering
\includegraphics[width=\linewidth]{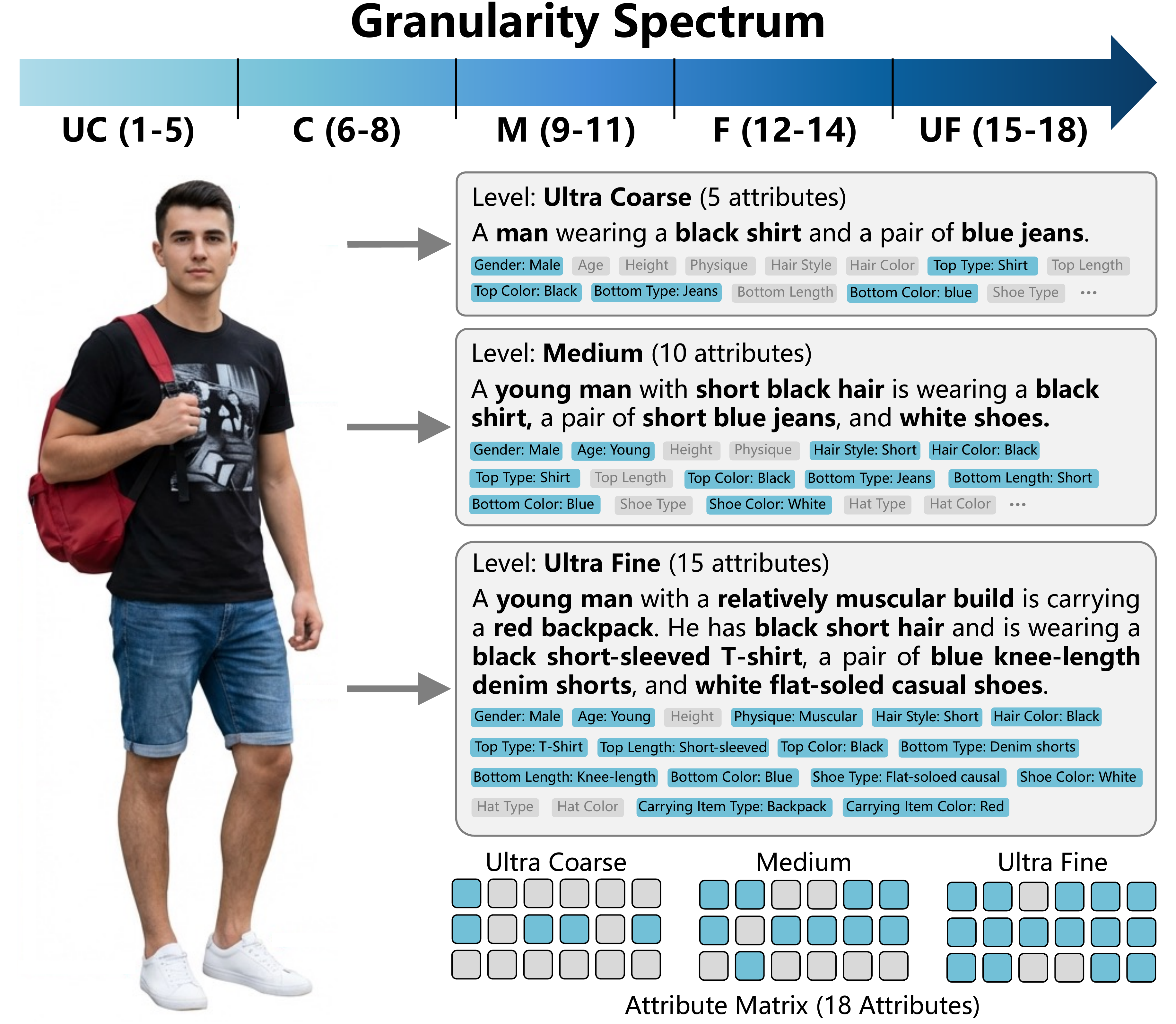}
\caption{The hierarchical taxonomy of textual granularity. Descriptions are systematically categorized into five levels (UC to UF) based on a unified ontology of 18 semantic attributes. This attribute-driven spectrum characterizes the informational density of person descriptions more accurately than text length.}
\label{fig:granularity_spectrum}
\vspace{-4mm}
\end{figure}

\section{DATA: Constructing Multi-grained Text-based Person Retrieval Dataset}

\textit{\textbf{Extension Note:} The conference paper contributed UFine6926, a single-granularity dataset with ultra-fine-grained annotations only. This section introduces three entirely new contributions: (1) a principled five-level granularity spectrum defined over 18 semantic attributes; (2) a Multi-grained Text Annotation Engine that leverages LLMs and VLMs for scalable granularity-controlled description generation; and (3) UFine6926-MG, the first dataset with balanced annotations across all five granularity levels.}

\subsection{Granularity Spectrum}
Accurately quantifying textual granularity is foundational to granularity-aware retrieval. Prevailing methods use text length (word or token count)~\cite{cuhkpedes,icfgpedes,rstpreid} as a proxy, which is fundamentally flawed: lengthy descriptions may be redundant, while concise ones can be highly informative. Granularity lies not in length, but in semantic informativeness. We therefore define granularity by the number of distinct semantic attributes a description conveys, and construct a unified ontology of 18 key attributes covering the full descriptive space, as shown in Fig.~\ref{fig:granularity_spectrum}: \textit{Gender, Skin Color, Age, Height, Physique, Hairstyle, Hair Color, Top Type, Top Length, Top Color, Bottom Type, Bottom Length, Bottom Color, Shoe Type, Hat Type, Hat Color, Carrying Item Type, and Carrying Item Color}. Building on this ontology, we introduce a five-level granularity taxonomy: Ultra Coarse-grained (UC): 1–5 attributes, Coarse-grained (C): 6–8, Medium-grained (M): 9–11, Fine-grained (F): 12–14, and Ultra Fine-grained (UF): 15–18. We acknowledge that both the ontology and taxonomy are empirically motivated: the 18 attributes span six semantically exhaustive groups — global biometrics, head region, upper body, lower body, footwear, and accessories — consistent with established ontologies in person ReID, while the five levels reflect natural stages of human descriptive behavior, whose practical sufficiency is validated by our experiments. A rigorous information-theoretic derivation of the optimal attribute set and level boundaries remains an open question for future work.

\subsection{Multi-grained Text Annotation Engine}
Building a multi-grained dataset requires textual descriptions spanning the full granularity spectrum, yet manually annotating each image at multiple granularity levels is prohibitively costly and unscalable. To address this, we propose a Multi-grained Text Annotation Engine that leverages large models~\cite{bai2025qwen2,yang2025qwen3} to automate the process, enabling an efficient and scalable pipeline for generating high-quality, granularity-controlled descriptions.
\begin{algorithm}[!h]
	\renewcommand{\algorithmicrequire}{\textbf{Input:}}
	\renewcommand{\algorithmicensure}{\textbf{Output:}}
	\caption{The pipeline of our annotation engine}
	\label{Alg1}
	\begin{algorithmic}[1]
	\REQUIRE The original image-text pair $\{I,T_0\}$, large language model $\mathcal{L}$, vision-language model $\mathcal{V}$, and human instructions $\mathcal{H} =\{h_a,h_r,h_f,h_e,h_t\}$;  
     \STATE Analyze the original text to obtain a set of not described attributes $S_{nda} = \mathcal{L}(T_0,h_a)$;
     \STATE Implement person attribute recognition to obtain supplementary descriptions $D_{supp}=\mathcal{V}(I,S_{nda},h_r)$;
     \STATE Integrate the text $T_0$ with supplementary descriptions to obtain an extended text $T_{ext}=\mathcal{L}(T_0,D_{supp},h_f)$;
     \STATE Extract the list of attribute descriptions in the extended text $D=\{d_k\}_{k=1}^N = \mathcal{L}(T_{ext},h_e)$;
     \IF{$N \leq N_c$}
      \STATE $T_{ext} \rightarrow{\mathcal{M}}$;
    \ELSE
    \STATE Initialize $n=N_c$, $D_{r} = D$, $D_s=\varnothing$;
    \WHILE{$n<N$ and $n<N_f$}
    \STATE $num \leftarrow N_{iv}$ \textbf{if} $n > N_c$ \textbf{else} $num \leftarrow N_c$
    \STATE $n_s = Random.int(num)$;
    \STATE $C_s = Random.sample(D_r,n_s)$;
    \STATE $D_s = D_s \cup C_s, D_r = D_s \setminus C_s$;
    \STATE $\mathbf{t}_m = \mathcal{L}(D_s,h_t), T_m \rightarrow{\mathcal{M}}$;
    \STATE $n=n+N_{iv}$;
    
    \ENDWHILE
     \ENDIF
        \ENSURE The set $\mathcal{M}$ containing multi-grained texts.
	\end{algorithmic} 
\label{Impl}
\end{algorithm}

Our engine follows a human-in-the-loop paradigm, bootstrapped from expert-annotated fine-grained texts~\cite{ufinebench}. It orchestrates a structured pipeline of transformations powered by role-specific LLMs~\cite{yang2025qwen3} and VLMs~\cite{bai2025qwen2} with carefully designed instructions. As outlined in Alg.~\ref{Impl}, the pipeline takes an image-text pair $(I, T_0)$ as input and applies five core procedures to produce a set of multi-grained texts $\mathcal{M}$:

\begin{figure*}[htb]
\centering
\includegraphics[width=0.98\linewidth]{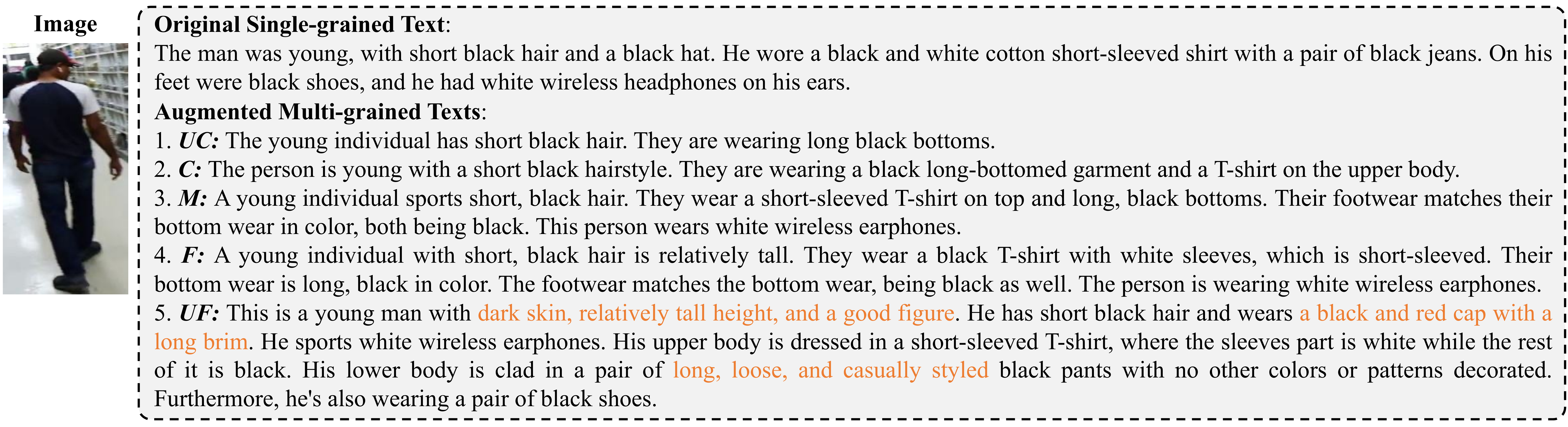}
\vspace{-1mm}
\caption{A representative sample of our UFine6926-MG. “UC”, “C”,
“M”, “F”, and “UF” denote ultra coarse-grained, coarse-grained, medium-grained, fine-grained and ultra fine-grained, respectively. The image details that are described in our ultra fine-grained text but not in the original text are highlighted in orange. Most images in UFine6926-MG are accompanied by five multi-grained texts of precise descriptions that range from ultra coarse to ultra fine detail.}
\label{fig:ufine6926mg}
\end{figure*}

1) Undescribed Attribute Analysis: An LLM $\mathcal{L}$ with instruction $h_a$ identifies attributes $S_{nda}$ from a predefined ontology that are absent in $T_0$. 2) Supplementary Description Generation: A VLM $\mathcal{V}$ with instruction $h_r$ generates supplementary descriptions $D_{supp}$ for attributes in $S_{nda}$ from image $I$. 3) Text-Attribute Integration: $\mathcal{L}$ with instruction $h_f$ merges $T_0$ and $D_{supp}$ into a coherent extended text $T_{ext}$. 4) Attribute Description Extraction: $\mathcal{L}$ with instruction $h_e$ parses $T_{ext}$ to extract a structured list $D = \{d_k\}_{k=1}^N$ of all $N$ attribute descriptions. 5) Multi-grained Text Generation: If $N \leq N_c$, $T_{ext}$ is directly included in $\mathcal{M}$; otherwise, a progressive strategy iteratively samples attribute subsets $D_s \subset D$ of increasing size and uses $\mathcal{L}$ to generate a fluent text $T_m$ per subset. Here $N_c$, $N_{iv}$, and $N_f$ are set to 5, 3, and 15, respectively, based on the granularity spectrum.

This progressive strategy naturally produces a query chain where each subsequent description contains additional attribute details, directly supporting the Progressive Granularity Assessment. By injecting precise human guidance via instructions $\mathcal{H}$ at each step, we ensure the accuracy and coherence of generated texts. This engine enables efficient construction of UFine6926-MG and MG-Eval evaluation queries, transforming multi-grained data acquisition from a manual burden into a scalable, automated process.

\vspace{-1mm}
\subsection{UFine6926-MG: A Multi-grained Dataset}

\begin{table}[t]
\centering
\caption{Quantitative analysis of training set granularity across datasets. Our UFine6926-MG provides a more diverse and equitable representation of multi-grained descriptions, effectively bridging the coverage gaps found in existing datasets.}
\label{granularity_statistics}
\resizebox{\linewidth}{!}{
\begin{tabular}{cccccc}
\toprule
 \multicolumn{1}{c}{\multirow{2}{*}{\textbf{Dataset}}}  & \multicolumn{5}{c}{\textbf{Granularity Statistics}} \\
  & UC & C & M & F&UF \\
\midrule  
CUHK-PEDES~\cite{cuhkpedes}&8,030 &35,639 &22,128&2,299 &30 \\
ICFG-PEDES~\cite{icfgpedes}&60&2,777&18,987&12,521&329\\ 
UFine6926~\cite{ufinebench}&8&85&1,584&14,064&21,413\\ 
\rowcolor{gray!20}
\textbf{UFine6926-MG}&37,154& 37,152& 37,140	&36,958 &34,219\\
\bottomrule
\end{tabular}
}
\vspace{-3mm}
\end{table}

We present UFine6926-MG, a multi-grained extension of UFine6926~\cite{ufinebench} constructed via a two-phase process to support any-granularity retrieval. 1) Base Dataset Construction. UFine6926 contains 26,206 images of 6,926 identities collected from diverse internet videos using FairMOT~\cite{zhang2021fairmot} and filtered via PLIP's strategies~\cite{zuo2024plip} and manual verification. Each image is annotated with two ultra fine-grained descriptions by 58 trained experts, averaging 80.8 words per description, significantly surpassing existing datasets in textual detail. 2) Multi-grained Augmentation. To achieve comprehensive granularity coverage, we automatically generate texts across all five granularity levels using our annotation engine. As shown in Tab.~\ref{granularity_statistics}, this rebalances the originally fine-grained-skewed UFine6926 into UFine6926-MG with evenly distributed texts across all levels (UC, C, M, F, UF). To ensure annotation quality, three trained annotators validated a randomly sampled subset of 100 identities for semantic accuracy and attribute completeness, yielding a pass rate of 94.7\%. Since all multi-grained texts are bootstrapped from the original expert-annotated ultra-fine descriptions, UF-level annotations are fully preserved, ensuring consistency between UFine6926-MG and UFine6926. A representative example is shown in Fig.~\ref{fig:ufine6926mg}. UFine6926-MG retains the original image set while providing balanced multi-grained textual coverage, establishing a foundational dataset for any-granularity person retrieval research.

\vspace{-1mm}
\section{EVALUATION: Benchmarking the Performance in Multi-grained Scenarios}

\textit{\textbf{Extension Note:} The conference paper proposed the UFine3C evaluation set and the mSD metric under a fixed ultra-fine-grained setting. This section introduces two new contributions: (1) MG-Eval, a multi-grained benchmark with cross-identity labels that formally models the one-to-many correspondence of coarse-grained queries; and (2) two tailored evaluation protocols—Separate and Progressive Granularity Assessment. The mSD metric is directly adopted here without modification.}

\subsection{Evaluation Set with Cross-Identity Labels}
We construct a novel evaluation set with cross-identity labels to better reflect real-world retrieval semantics where coarse queries legitimately match multiple identities. This benchmark comprises 1,000 identities curated from SYSU-MM01~\cite{wu2017rgb} and LLCM~\cite{zhang2023diverse} datasets, ensuring domain diversity for rigorous generalization testing. For each identity, we manually author an ultra fine-grained description and use our Multi-grained Text Annotation Engine to generate four additional descriptions across granularities. The key innovation is assigning cross-identity labels through attribute matching: coarse-grained queries are linked to all identities whose fine-grained attributes semantically encompass them. This process combines LLM-assisted annotation~\cite{yang2025qwen3} with human verification for accuracy. The resulting set contains 5,000 textual queries (balanced across five granularities) and 45,113 gallery images, where each query may have multiple valid matches. This enables more realistic evaluation of retrieval performance in multi-grained scenarios.

\subsection{Evaluation Protocols}
\textbf{Separate Granularity Assessment.}
This protocol evaluates a model’s ability to retrieve accurate ranking lists from a gallery when queried with texts of varying granularity. Let the granularity space be divided into $M$ categories, and let $N_i$ denote the number of text queries belonging to the $i$-th granularity category. If $ps_{i,j}$ represents the performance score of metric $j$ for queries of granularity $i$, the overall multi-grained performance for metric $j$ is computed as the weighted average:
\begin{equation}
metric_j = \sum_{i=1}^{M} \frac{{N_i}\cdot{ps_{i,j}}}{\sum_{i=1}^{M} N_i}.
\end{equation}
Following standard practice in text-based person retrieval~\cite{cuhkpedes}, we adopt the popular rank-\textit{k} (\textit{k}=1,5,10) and mAP as the underlying metrics for assessing ranking quality.

\begin{figure}[htb]
\centering
\includegraphics[width=\linewidth]{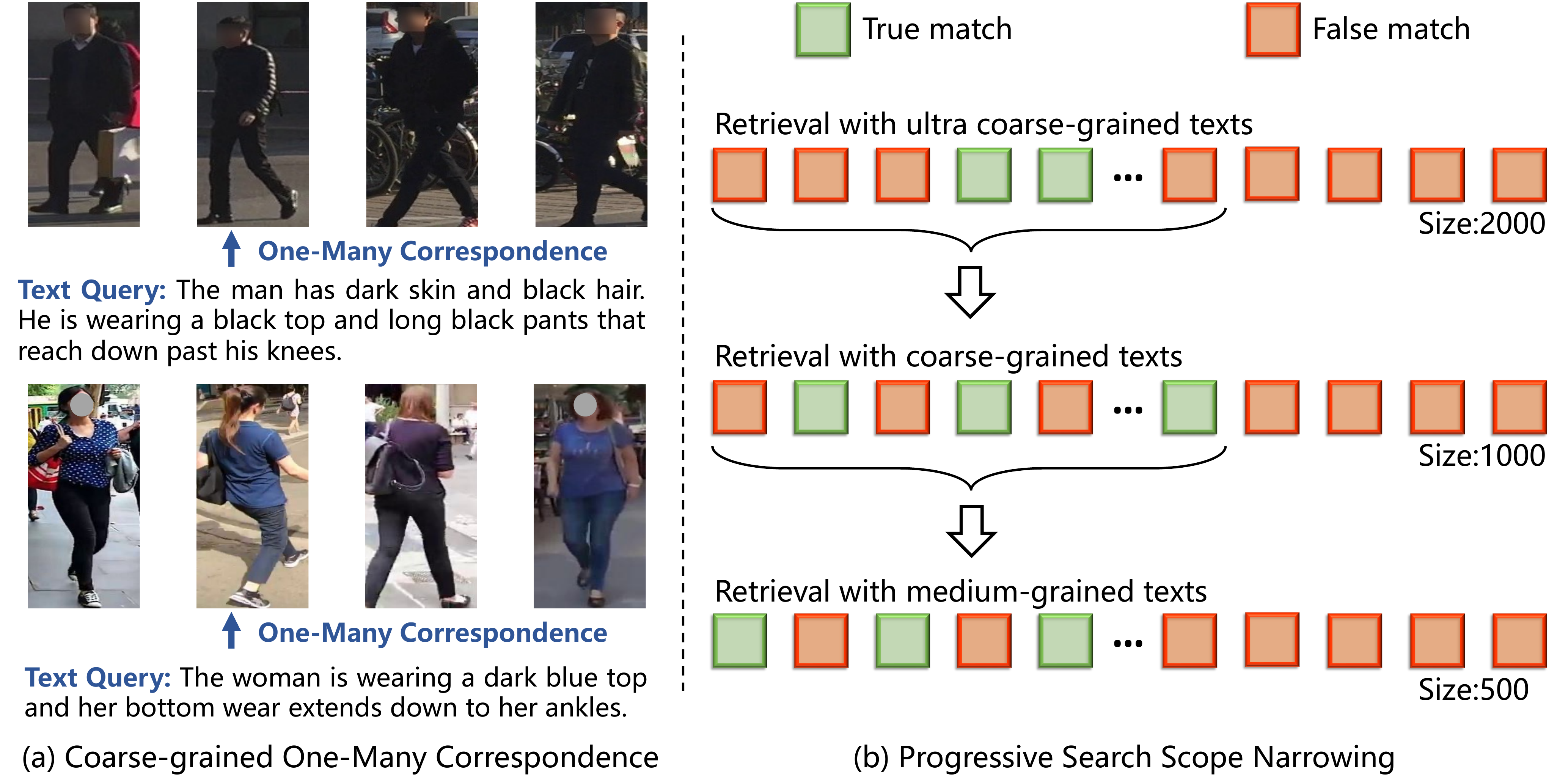}
\caption{Motivation for progressive-granularity assessment. (a) One-to-many correspondence inherent in coarse-grained retrieval. (b) The progressive shrinking of the gallery pool as query specificity increases during the retrieval process.}
\label{fig:pge}
\end{figure}
\noindent
\textbf{Progressive-Granularity Assessment.}
Coarse-grained textual queries, being information-sparse, often correspond correctly to multiple person identities in real-world settings. This one-to-many matching behavior is not adequately captured by conventional evaluation protocols, which rely on binary identity-based relevance judgments. To further reflect retrieval performance under realistic multi-grained query settings, we propose a Progressive Granularity Assessment protocol.

As illustrated in Fig.~\ref{fig:pge} (b), the core idea is to iteratively refine the candidate gallery using a sequence of increasingly detailed textual descriptions~\cite{bai2025chat,lu2025llavareid}. Formally, given a retrieval instance with $n$ progressively refined textual queries $\{t^j\}_{j=0}^{n-1}$, starting from the original gallery $G^0$, the gallery is recursively filtered as:
\begin{equation}
\begin{cases}
k^j = \rho \cdot{|G^j|} \\
G^{j+1} = \mathbf{Top}k^j\{\mathbf{Rank}(t^j,G^j)\},
\end{cases}
\end{equation}
where $\mathbf{Rank}(t^j,G^j)$ sorts gallery samples by similarity to query $t^j$, $\mathbf{Top}k^j(\cdot)$ selects the top $k^j$ candidates, $|G^j|$ denotes gallery size at step $j$, and $\rho$ is a reduction rate. The final ranking list is obtained by computing similarity between the finest-grained query $t^{n-1}$ and the last-stage gallery $G^{n-1}$, offering a more realistic assessment of retrieval capability under granularity-varying conditions.

\subsection{A New Evaluation Metric}
\begin{figure}[htb]
\centering
\vspace{-2mm}
\includegraphics[width=0.8\linewidth]{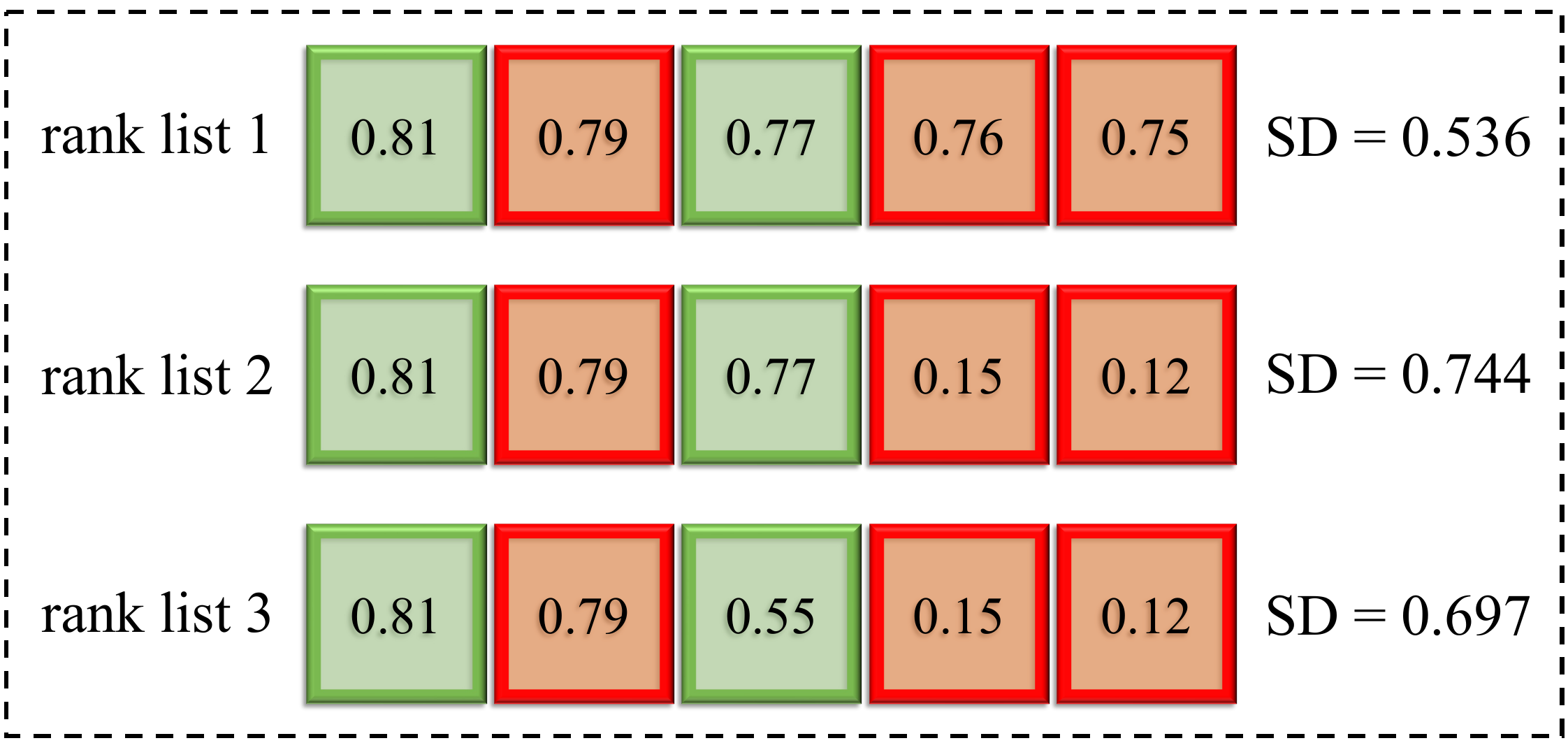}
\caption{Sensitivity comparison of SD vs. AP. Green and red boxes represent correct and incorrect matches. AP is identical (0.833) for all three cases, whereas SD (0.536, 0.744, and 0.697) successfully distinguishes between different ranking distributions, highlighting its superior discriminative power.}
\label{fig:mapweak}
\end{figure}

While mean Average Precision (mAP) is widely adopted in person retrieval benchmarks~\cite{cuhkpedes,icfgpedes,rstpreid}, it has inherent limitations in capturing fine-grained performance differences. As a rank-based metric, mAP relies on discrete ranking positions, ignoring continuous similarity values that better reflect retrieval capability. As illustrated in Fig.~\ref{fig:mapweak}, three ranking lists with significantly different similarity distributions can yield identical AP scores of 0.833, revealing the metric's inability to distinguish qualitatively different retrieval outcomes.

To address this gap, we propose the mean Similarity Distribution (mSD) metric, which operates directly on continuous similarity values for more sensitive retrieval evaluation. As shown in Fig.~\ref{fig:mapweak}, mSD successfully discriminates between the three ranking lists with scores of 0.536, 0.744, and 0.697, accurately reflecting their qualitative differences.

Formally, given a ranked list of $n$ samples $\{s_i\}_{i=1}^n$ with $s_i \in [0,1]$, where $s^+$ and $s^-$ denote similarity values of matched and unmatched samples, mSD is computed as follows.

First, the Precision-Normalized Ratio (PNR) measures the normalized similarity contrast between positive and negative samples:
\begin{equation}
    \text{PNR} = 1 - e^{-k \cdot x}, \quad \text{where } x = \frac{\mathbb{E}[s^+]}{\mathbb{E}[s^-]},
\end{equation}
where $\mathbb{E}[\cdot]$ is the arithmetic mean, $k=1$ is a scaling factor, and $x$ is the ratio of average matched to unmatched similarity.

Next, the Average Similarity Precision (ASP) measures the cumulative similarity contribution of relevant samples throughout the ranking:
\begin{equation}
    \text{ASP} = \frac{1}{n^+} \sum_{k=1}^{n^+} \frac{\sum_{i=1}^{j_k} s_i^+}{\sum_{i=1}^{j_k} s_i},
\end{equation}
where $\{j_k\}_{k=1}^{n^+}$ denotes the rank positions of the $n^+$ matched samples.

The Similarity Distribution (SD) for a single query is then:
\begin{equation}
    \text{SD} = \text{PNR} \times \text{ASP}.
\end{equation}

Finally, \textbf{mSD} is obtained by averaging SD over all queries. By jointly capturing ranking quality and similarity distribution, mSD is particularly suited for evaluating fine-grained retrieval performance in multi-grained scenarios.

\begin{figure*}[htb]
\centering
\includegraphics[width=\linewidth]{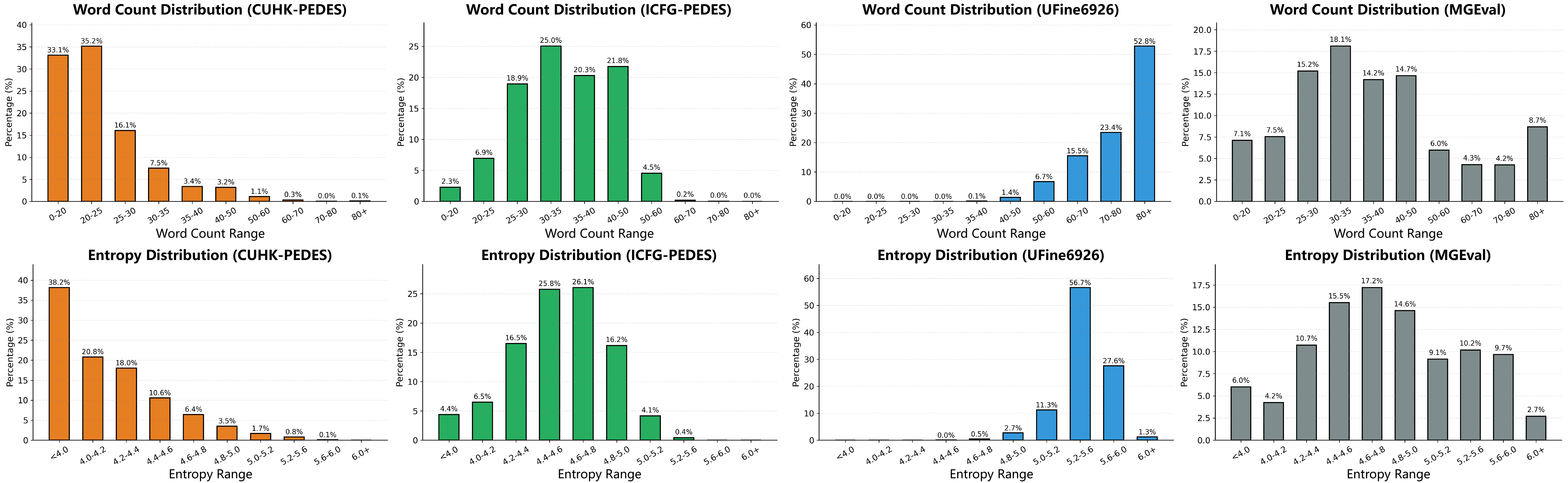}
\caption{Granularity distributions across different benchmarks. Rows represent word count (length) and information entropy (richness) densities, respectively. Columns (left to right): CUHK-PEDES~\cite{cuhkpedes}, ICFG-PEDES~\cite{icfgpedes}, UFine6926~\cite{ufinebench}, and MG-Eval. While existing benchmarks exhibit narrow coverage, MG-Eval achieves a more uniform spectrum across both metrics.}
\label{fig:Narrow_Granularity}
\end{figure*}

\section{See Both Forest and Trees? Systematic Diagnosing for Existing Research}

\textit{\textbf{Extension Note:} This section is entirely new with no counterpart in the conference paper. We conduct a systematic diagnosis from both data-centric and method-centric perspectives.}

This section systematically diagnoses limitations in current person retrieval research through the lens of granularity adaptation. The ``forest and trees'' metaphor encapsulates the core challenge: models must simultaneously recognize coarse-grained contexts (``forest'') and fine-grained details (``trees''). Our analysis reveals fundamental deficiencies in existing research across both data composition and method design, demonstrating their inability to handle real-world granularity variations. These findings collectively motivate the solutions we introduce subsequently.

\subsection{Data-Centric Diagnosis}

\noindent
\textbf{Narrow Granularity Spectrum of Existing Benchmarks.}
We systematically analyze the granularity distributions of major benchmarks~\cite{cuhkpedes,icfgpedes,ufinebench} via word count and information entropy. As shown in Fig.~\ref{fig:Narrow_Granularity}, existing datasets cover only fragmented segments of the granularity spectrum: CUHK-PEDES~\cite{cuhkpedes} concentrates in the low-granularity regime, ICFG-PEDES~\cite{icfgpedes} clusters in the intermediate range, and UFine6926~\cite{ufinebench} skews heavily toward the fine-grained end. In contrast, MG-Eval exhibits a significantly more balanced and uniform distribution across both metrics, bridging these coverage gaps and providing a comprehensive platform that captures the full diversity of query specificities in real-world retrieval scenarios.

\begin{figure}[htb]
\centering
\vspace{-2mm}
\includegraphics[width=\linewidth]{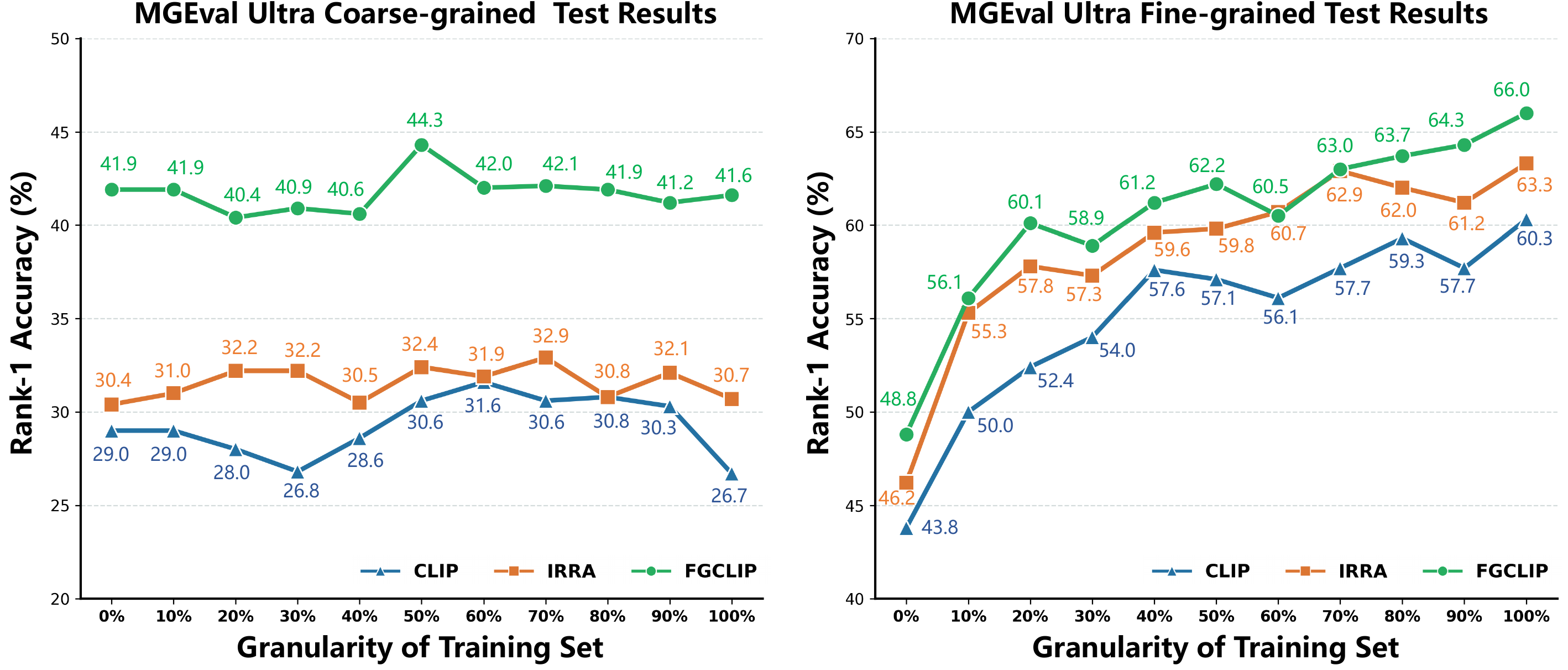}
\caption{The pervasive effect of granularity bias on retrieval performance~\cite{clip,irra,xie2025fgclip}. We evaluate the rank-1 accuracy on ultra coarse-grained and ultra fine-grained test sets by varying the proportion of high-detail samples in the training data.}
\label{fig:granularity_bias}
\end{figure}

\noindent
\textbf{The Pervasive Effect of Granularity Bias.} To systematically quantify how training data distribution affects performance across granularities, we conduct a controlled experiment by varying the ratio of low-detail (UC and C) and high-detail (F and UF) texts while keeping the total number of training samples constant. As illustrated in Fig.~\ref{fig:granularity_bias}, the results reveal a distinct asymmetric sensitivity to training granularity: for the ultra-fine test set, rank-1 accuracy exhibits a significant and monotonic improvement as the proportion of high-detail training data increases. Conversely, performance on the ultra-coarse test set remains relatively stagnant, showing marginal gains regardless of the increased training granularity. This divergence suggests that while high-detail data is essential for learning discriminative features, its increased information density does not inherently resolve the semantic ambiguity inherent in coarse-grained queries. These findings confirm that performance bias is driven by training data composition, underscoring the necessity of a balanced granularity spectrum for robust retrieval.

\begin{figure}[htb]
\centering
\vspace{-3mm}
\includegraphics[width=\linewidth]{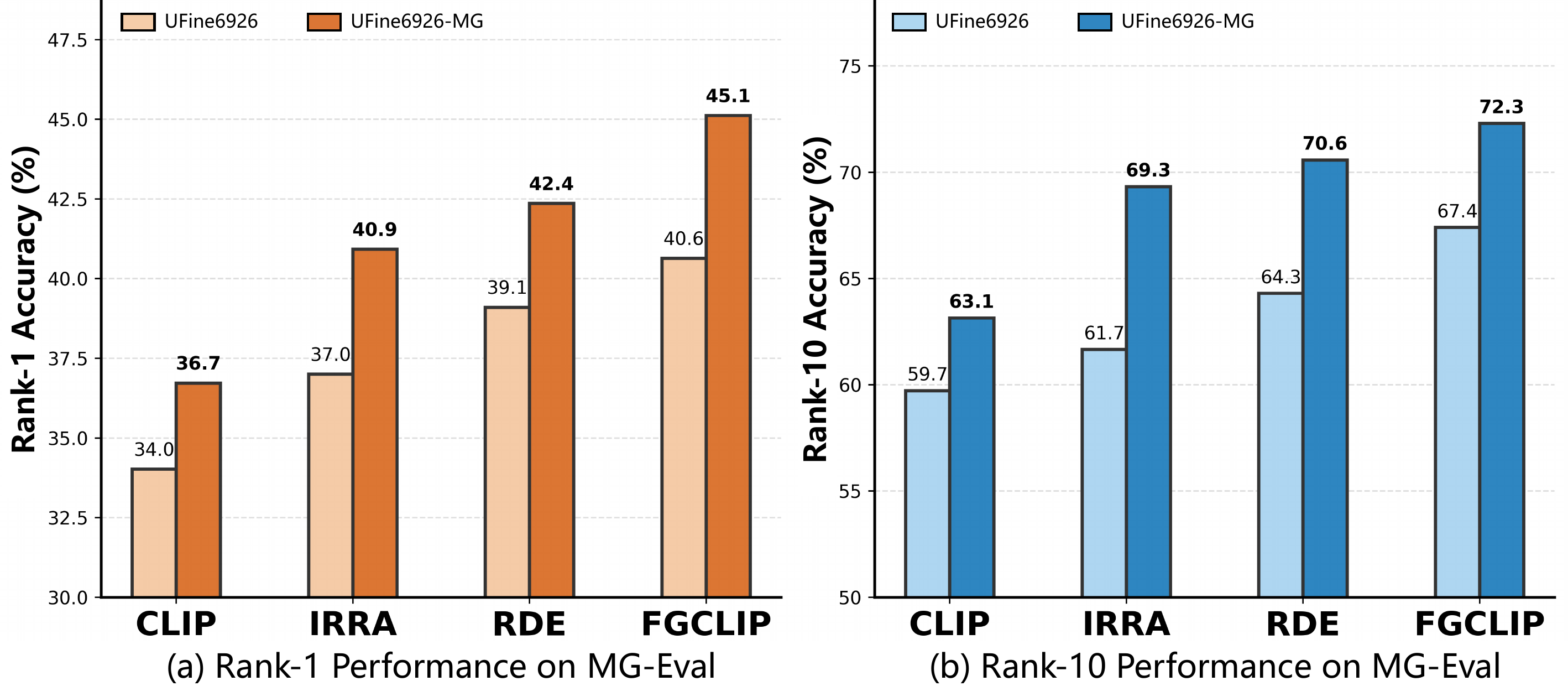}
\caption{Impact of balanced granularity distribution. The consistent gains across models trained on UFine6926-MG underscore the necessity of a balanced granularity spectrum. }
\label{fig:Balanced Granularity Distribution}
\end{figure}

\noindent
\textbf{The Advantage of a Balanced Granularity Spectrum.} To evaluate the impact of training data distribution, we compare the performance of four representative models~\cite{clip,irra,rde,xie2025fgclip} trained on either the original UFine6926~\cite{ufinebench} or our granularity-balanced UFine6926-MG. As illustrated in Fig.~\ref{fig:Balanced Granularity Distribution}, training on the balanced UFine6926-MG leads to a consistent and significant improvement in the separate granularity assessment across the MG-Eval benchmark for all evaluated architectures. This gain demonstrates that the balanced distribution effectively rectifies the extreme data bias inherent in the original fine-grained dataset. By providing a more uniform exposure to varying levels of descriptive detail during training, UFine6926-MG enables models to learn a more robust and comprehensive semantic space, ultimately yielding superior generalizability across the entire granularity spectrum.
\begin{figure}[htb]
\centering
\vspace{-2mm}
\includegraphics[width=\linewidth]{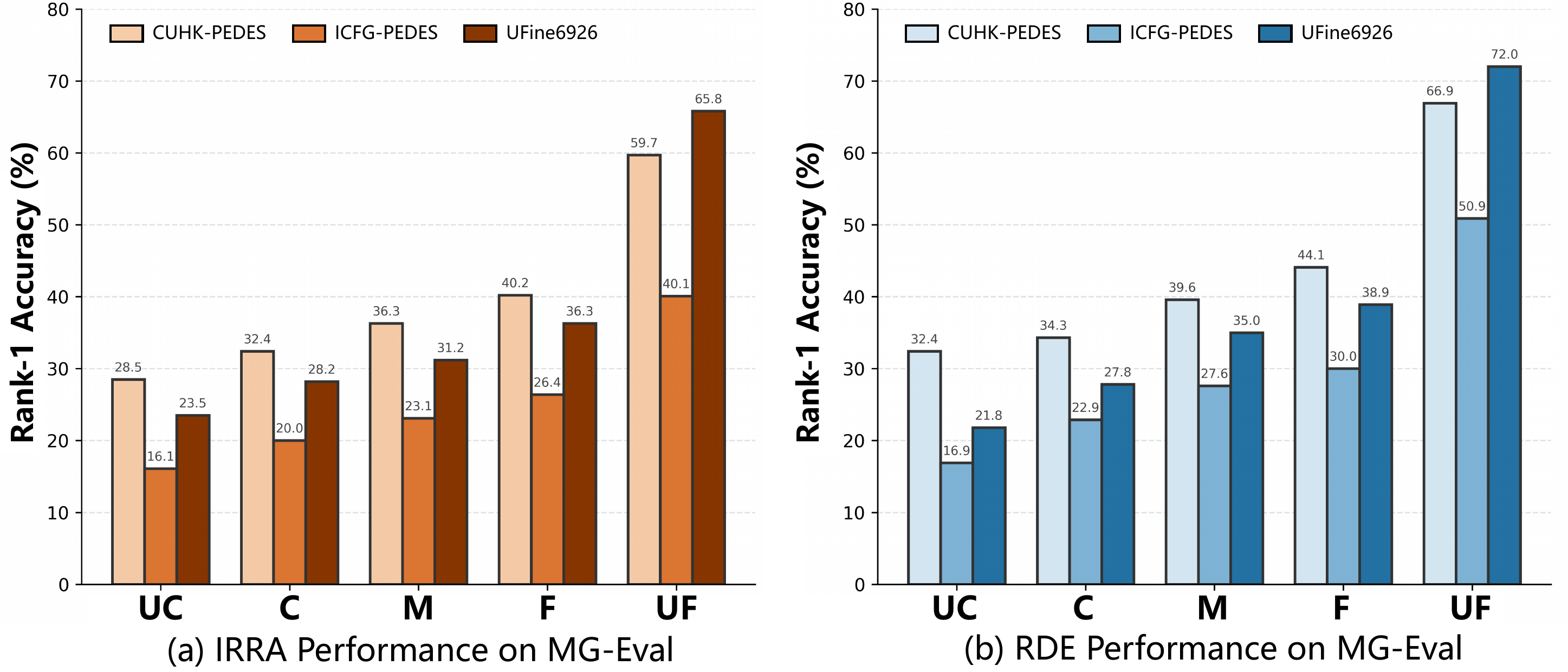}
\vspace{-6mm}
\caption{Performance sensitivity across granularity levels. For any given combination of method and training dataset, performance improves consistently as granularity increases. However, the significant performance fluctuations across different levels result in high variance.}
\vspace{-3mm}
\label{fig:Granularity_Sensitivity}
\end{figure}

\subsection{Method-Centric Diagnosis}
\noindent
\textbf{Revealing Granularity Sensitivity.} 
We conduct a granularity sensitivity analysis to assess how existing models handle varying text granularity. Two representative models, IRRA~\cite{irra} and RDE~\cite{rde}, are trained on three standard datasets~\cite{cuhkpedes,icfgpedes,ufinebench} and evaluated on MG-Eval across all granularity levels. Results (Fig.~\ref{fig:Granularity_Sensitivity}) reveal significant performance variance across granularities for all model-dataset combinations. While performance generally improves with finer granularity, substantial performance fluctuations are observed across the granularity spectrum. This sensitivity pattern persists regardless of the training dataset, indicating a fundamental limitation in current methods. These findings demonstrate that existing retrieval models lack robustness to granularity variation. Their performance becomes unpredictably dependent on query granularity, highlighting the critical need for granularity-adaptive approaches in real-world scenarios where query specificity naturally varies.

\begin{figure}[htb]
\centering
\includegraphics[width=\linewidth]{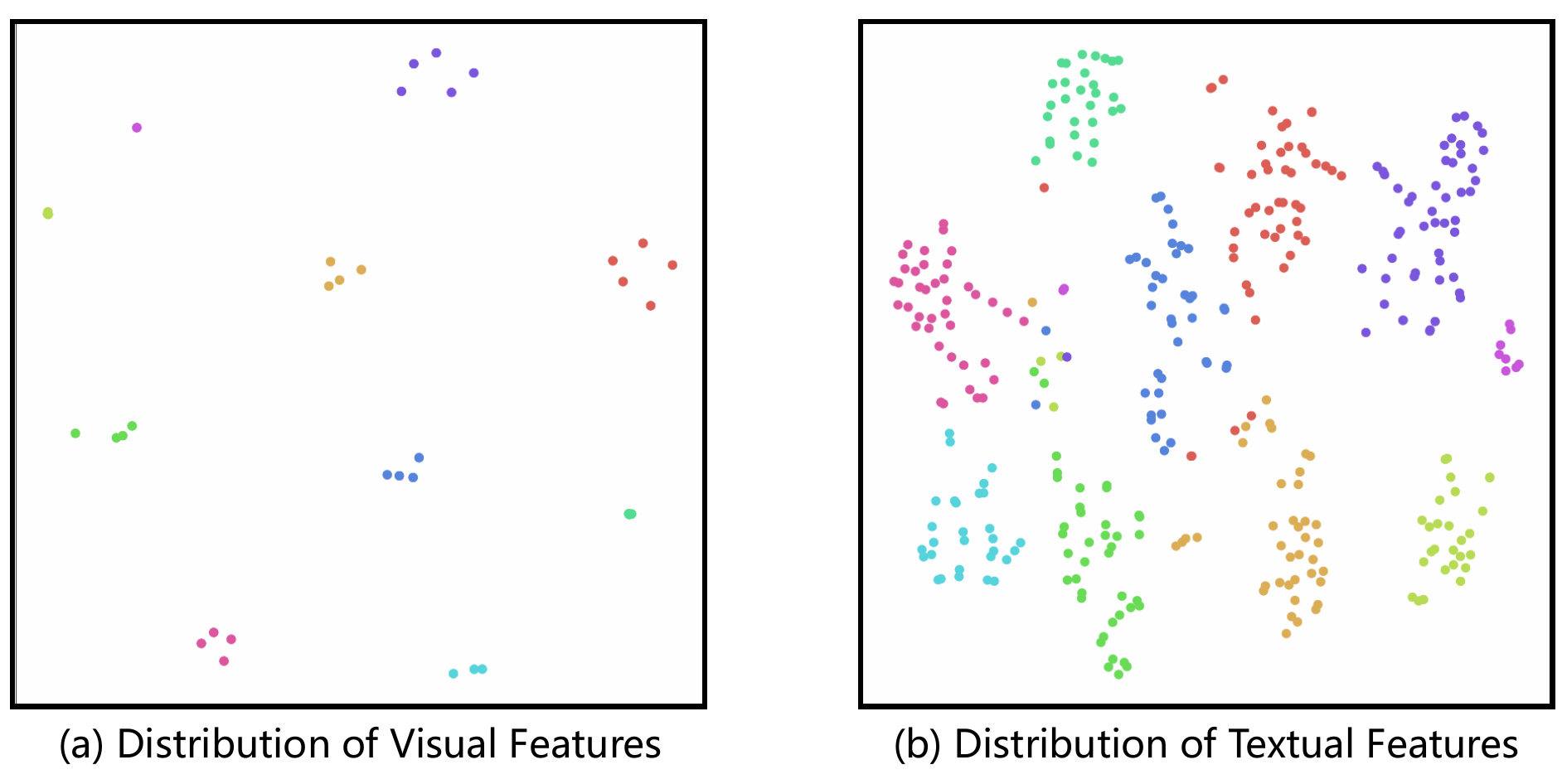}
\vspace{-6mm}
\caption{T-SNE visualization of CLIP-extracted features for 10 random identities. Visual features form tight, localized clusters with low uncertainty (uni-granular), whereas textual features are markedly more scattered, reflecting the inherent semantic uncertainty of their multi-granular representations.}
\vspace{-1mm}
\label{fig:Single-Granularity Bottleneck}
\end{figure}
\noindent
\textbf{The Single-Granularity Bottleneck in Visual Encoding.} 
A fundamental limitation of existing methods lies in the granularity mismatch between visual and textual representations. While a single image is typically encoded into a unique, single-grained feature vector, it is often paired with multiple textual descriptions of varying granularities in multi-grained datasets. To illustrate this discrepancy, we visualize CLIP~\cite{clip} features of 10 random identities from UFine6926-MG using t-SNE (Fig.~\ref{fig:Single-Granularity Bottleneck}). The results clearly show that visual features form compact clusters, whereas their corresponding multi-grained textual features are widely dispersed in the embedding space. This demonstrates that while visual encoders produce single-grained representations, textual features inherently capture multi-grained semantic information. This granularity mismatch introduces significant optimization challenges during training. Forcing alignment between a single-grained image embedding and multiple text embeddings of different granularities creates conflicting learning signals, undermining the model's ability to learn stable cross-modal representations.

\noindent
\textbf{The Crucial Gap: Granularity-Aware Fuzzy Alignment.}
Our analysis reveals a fundamental limitation in existing methods: their inability to adequately model the fuzzy alignment characteristics inherent in multi-grained scenarios. This limitation manifests through two distinct aspects.

1. The Rigid Binary Matching Paradigm. Current methods operate under a binary distinction that fails to capture the nuanced nature of multi-grained contexts. The widely adopted contrastive learning objective, as exemplified by CLIP-style models~\cite{clip,irra,rde}, follows the formulation:
\begin{equation}
\mathcal{L}_{\text{cont}} = -\frac{1}{N}\sum_{i=1}^N \log\frac{\exp(s(I_i,T_i)/\tau)}{\sum_{j=1}^N \exp(s(I_i,T_j)/\tau)}.
\end{equation}
This formulation inherently treats all non-identical pairs as negative samples, regardless of their potential semantic relevance. The gradient analysis reveals the core issue:
\begin{equation}
\frac{\partial\mathcal{L}_{\text{cont}}}{\partial s(I_i,T_j)} = \begin{cases}
\frac{1}{\tau}(p_{ij}-1) & \text{for positive pairs } (i=j) \\
\frac{1}{\tau}p_{ij} & \text{for negative pairs } (i\neq j),
\end{cases}
\end{equation}
where $p_{ij}$ represents the softmax probability. This gradient forces all non-matching pairs toward low similarities, thereby penalizing semantically relevant but non-identical matches that frequently occur in coarse-grained scenarios.

2. The Uniform Similarity Fallacy. Current methods enforce uniform similarity targets for all positive pairs, ignoring granularity-induced variations in match strength. From an information-theoretic perspective, optimal similarity should reflect the discriminative information in textual descriptions. Formally, for fine-grained text $T_f$ and coarse-grained text $T_c$ describing the same image $I$:
\begin{equation}
s^*(I, T_f) - s^*(I, T_c) = I(I;A_f \backslash A_c | A_c) > 0,
\end{equation}
here $I(\cdot;\cdot)$ denotes mutual information, with $A_f$ and $A_c$ ($A_c \subset A_f$) representing attribute sets for fine- and coarse-grained texts respectively. This conditional mutual information measures the additional discriminative power of fine-grained attributes and is strictly positive when they provide meaningful supplemental information.

\begin{figure*}[htb]
\centering
\includegraphics[width=\linewidth]{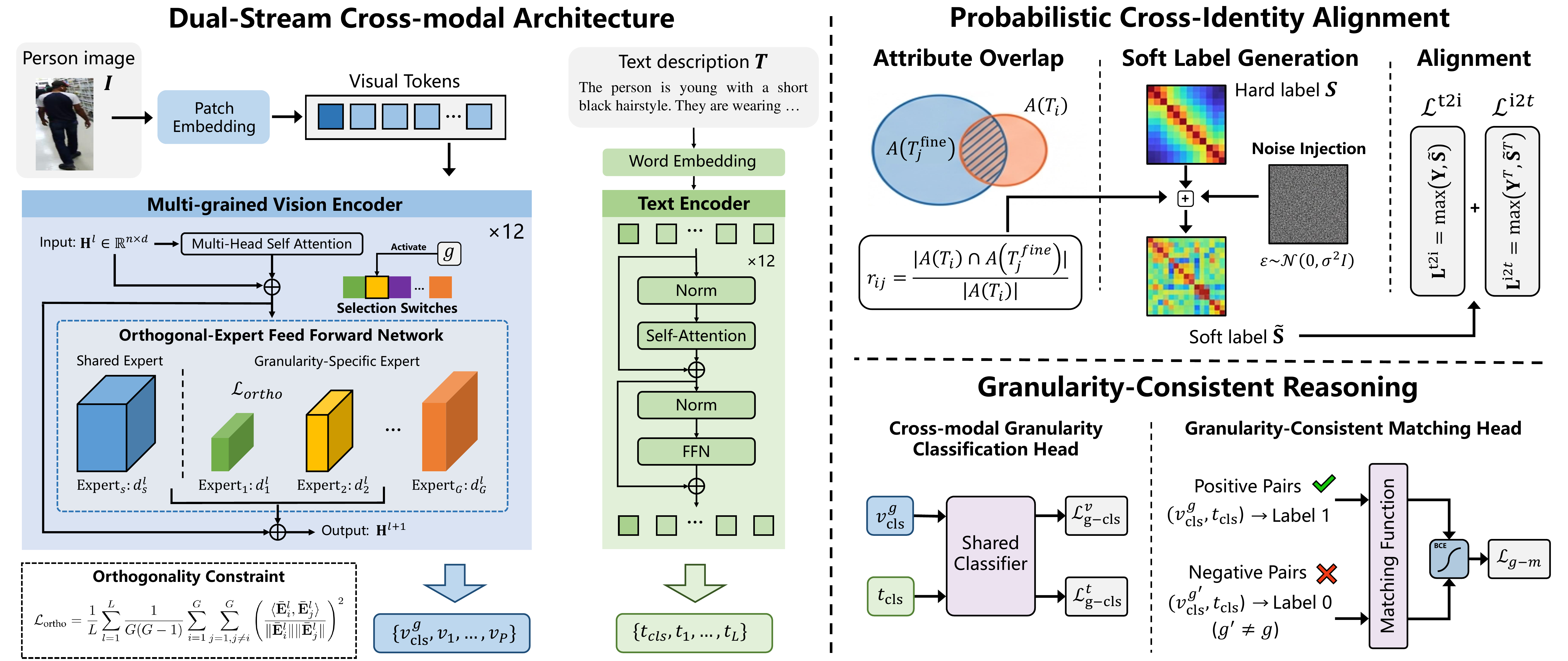}
\vspace{-5mm}
\caption{Overview of the proposed CMAM framework. It consists of three synergistic modules: (1) Orthogonal-Expert Multi-grained Visual Perception (OE-MVP), which employs structurally heterogeneous experts with progressive capacities, regulated by an orthogonality constraint to extract disentangled visual features; (2) Probabilistic Cross-Identity Alignment (PCIA), which addresses the inherent ambiguity of coarse-grained queries by generating soft labels based on attribute overlap and Gaussian noise injection; and (3) Granularity-Consistent Reasoning (GCR), which utilizes a dual-head structure for granularity classification and binary matching to ensure semantic-granularity consistency across the entire spectrum.}
\vspace{-3mm}
\label{fig:Method}
\end{figure*}
The prevailing constraint in contemporary contrastive learning frameworks:
\begin{equation}
s(I,T) \to \gamma_{\text{high}}, \quad \forall (I,T) \in \mathcal{P},
\end{equation}
where $\mathcal{P}$ denotes the set of positive pairs, fundamentally violates this optimal similarity structure. This uniform optimization strategy prevents models from learning the granularity-dependent calibration. 

These limitations lead to models that both overly reject valid coarse-grained matches and poorly distinguish confidence levels across matches. This necessitates a shift to granularity-aware fuzzy alignment to accommodate the continuous spectrum of real-world matching relationships.

\section{ALGORITHM: Cross-modal Multi-grained Aligning and Matching Framework}

\textit{\textbf{Extension Note:} The conference paper proposed CFAM based on a shared granularity decoder and hard negative matching. CMAM is a fundamental redesign with three entirely new modules: (1) Orthogonal-Expert Multi-grained Visual Perception for granularity-disentangled feature learning; (2) Probabilistic Cross-Identity Alignment for modeling many-to-many coarse-grained matching via soft labels; and (3) Granularity-Consistent Reasoning for explicit cross-modal granularity supervision. None of these modules has a counterpart in CFAM.}

\subsection{Architecture}
As shown in Fig.~\ref{fig:Method}, our CMAM framework is built upon a dual-stream architecture featuring a specially designed multi-grained visual encoder. While the textual encoder follows a standard Transformer~\cite{vaswani2017attention} design, the visual encoder is enhanced with granularity-aware components to dynamically adapt to varying levels of textual specificity.

Formally, the model takes an image-text pair $(I, T)$ with identity label $y \in \{1, \dots, C\}$ and granularity label $g \in \{1, \dots, G\}$ as input. Image $I$ is divided into $P$ non-overlapping patches and fed into the multi-grained visual encoder (with parameters selectively activated based on $g$), producing visual representations $\{\mathbf{v}_{\text{cls}}, \mathbf{v}_1, \dots, \mathbf{v}_P\}$. Text $T$ is trimmed to a uniform length $L_T$, tokenized, and fed into the textual encoder, producing textual representations $\{\mathbf{t}_{\text{cls}}, \mathbf{t}_1, \dots, \mathbf{t}_{L_T}\}$. Here, $\mathbf{v}_{\text{cls}}, \mathbf{t}_{\text{cls}} \in \mathbb{R}^d$ serve as the global representations for the image and text, respectively.

The following sections detail our core technical contributions. Sec.~\ref{sec6.2} introduces the Orthogonal-Expert Multi-grained Visual Perception module, which uses structurally heterogeneous experts with orthogonality-constrained shared and specific pathways to extract discriminative features across granularities. Sec.~\ref{sec6.3} presents the Probabilistic Cross-Identity Alignment mechanism, which resolves coarse-grained matching ambiguity via attribute-overlap-based soft labels and injected noise. Sec.~\ref{sec6.4} describes the Granularity-Consistent Reasoning module, which verifies granularity consistency to provide fine-grained supervisory signals. Finally, Sec.~\ref{sec6.5} outlines the joint optimization strategy that unifies these components into a coherent framework.

\subsection{Orthogonal-Expert Multi-grained Visual Perception}
\label{sec6.2}
To enable granularity-aware visual representation learning, we fundamentally redesign the Feed-Forward Network (FFN) in each Transformer block of the vision encoder. Traditional ViTs employ a homogeneous MLP structure across all inputs, which lacks the adaptability to handle varying textual granularities. Our solution replaces the standard single MLP with a composite structure comprising multiple granularity-specific experts and one granularity-shared expert, allowing the model to dynamically adjust its feature extraction strategy based on the input granularity level.

Formally, for the $l$-th Transformer block with input feature $\mathbf{H}^l \in \mathbb{R}^{n \times d}$, the enhanced FFN operates as:
\begin{equation}
    \text{FFN}^l(\mathbf{H}^l) = \frac{1}{2} \left( \text{FFN}_{\text{shared}}^l(\mathbf{H}^l) + \text{FFN}_{g}^l(\mathbf{H}^l) \right),
\end{equation}
where $\text{FFN}_{\text{shared}}^l$ denotes the granularity-shared expert that remains active for all inputs to capture universal visual patterns, while $\text{FFN}_{g}^l$ represents one of $G$ granularity-specific experts selectively activated based on the granularity label $g$. This discrete expert selection enables specialized feature extraction aligned with textual specificity—coarser queries activate experts optimized for global attributes, while finer queries engage experts specialized in local details.

A key innovation lies in our \textit{structurally heterogeneous expert design}. Rather than using identical architectures, we deliberately vary the hidden dimensions across experts to align with their granularity roles. Formally, for expert $g$ in layer $l$, the two linear transformations are defined as:
\begin{equation}
    \text{FFN}_{g}^l(\mathbf{H}^l) = \sigma_{act}(\mathbf{H}^l \mathbf{W}_{g,1}^l + \mathbf{b}_{g,1}^l) \mathbf{W}_{g,2}^l + \mathbf{b}_{g,2}^l,
\end{equation}
where $\sigma_{act}$ denotes the activation function, $\mathbf{W}_{g,1}^l \in \mathbb{R}^{d \times d_g^l}$ and $\mathbf{W}_{g,2}^l \in \mathbb{R}^{d_g^l \times d}$, the intermediate dimension $d_g^l$ varies across granularities following $d_1^l < d_2^l < \cdots < d_G^l$. This progressive expansion of capacity from coarse to fine experts naturally accommodates the increasing complexity of visual details required at finer granularities.

To further ensure diversity and minimize redundancy among experts, we introduce an \textit{orthogonality constraint}. For each layer $l$, given the outputs of all granularity-specific experts $\{\mathbf{E}_1^l, \mathbf{E}_2^l, \dots, \mathbf{E}_G^l\}$ where $\mathbf{E}_g^l = \text{FFN}_{g}^l(\mathbf{H}^l)$, we compute the orthogonality loss as:
\begin{equation}
    \mathcal{L}_{\text{ortho}} = \frac{1}{L} \sum_{l=1}^{L} \frac{1}{G(G-1)} \sum_{i=1}^{G} \sum_{j=1, j \neq i}^{G} \left( \frac{\langle \bar{\mathbf{E}}_i^l, \bar{\mathbf{E}}_j^l \rangle}{\|\bar{\mathbf{E}}_i^l\| \|\bar{\mathbf{E}}_j^l\|} \right)^2,
\end{equation}
where $\bar{\mathbf{E}}_g^l \in \mathbb{R}^d$ denotes the mean-pooled representation of expert output $\mathbf{E}_g^l$ along the sequence dimension, and $L$ is the total number of layers. This formulation penalizes high correlations between different experts' outputs, encouraging each expert to develop specialized, non-redundant representations for its designated granularity level.

\textbf{Discussion.} The Orthogonal-Expert module establishes a theoretically grounded framework for visual feature disentanglement through structured subspace decomposition. By formulating the feature space as \(\mathcal{V} = \mathcal{V}_{\text{shared}} \oplus \mathcal{V}_1 \oplus \cdots \oplus \mathcal{V}_G\) with orthogonality constraints \(\mathcal{V}_i \perp \mathcal{V}_j\), we mathematically guarantee minimal interference between granularity-specific representations. The orthogonality loss \(\mathcal{L}_{\text{ortho}}\) implements this constraint by minimizing the cosine similarity between expert outputs, effectively enforcing feature diversity. Furthermore, the capacity progression \(d_1^l < \cdots < d_G^l\) embodies an information-theoretic principle: finer granularities necessitate higher-dimensional representations to encode increased visual complexity. This hierarchical capacity allocation serves as an implicit regularizer, preventing overfitting in coarse-grained experts while ensuring sufficient representational power for fine-grained details.

\subsection{Probabilistic Cross-Identity Alignment}
\label{sec6.3}
To address the many-to-many matching semantics inherent in coarse-grained 
queries~\cite{ufinebench}, we propose a Probabilistic Cross-Identity Alignment 
(PCIA) mechanism that explicitly models matching ambiguity via attribute-based 
soft labeling and noise injection, in contrast to traditional methods~\cite{clip,irra} 
that enforce strict one-to-one identity matching.

Formally, for a coarse-grained text query $T_i$ with granularity 
$g_i \leq g_{\text{coarse}}$, we extract its attribute set $\mathcal{A}(T_i)$ 
along with $\mathcal{A}(T_i^{\text{fine}})$ from the finest-grained description 
$T_i^{\text{fine}}$ of image $I_i$. For each other identity $j \neq i$ in the 
training batch, we compute the attribute overlap ratio between $T_i$ and the 
finest-grained text $T_j^{\text{fine}}$:
\begin{equation}
r_{ij} = \frac{|\mathcal{A}(T_i) \cap \mathcal{A}(T_j^{\text{fine}})|}{|\mathcal{A}(T_i)|}.
\end{equation}

We construct the soft similarity matrix $\mathbf{S}$, where $S_{ij} = r_{ij}$ 
if $r_{ij} \geq \theta$ and 0 otherwise, with threshold $\theta$ filtering out 
semantically trivial overlaps. To simulate real-world ambiguity, we inject Gaussian noise, where $\sigma$ controls the noise intensity and $\text{clip}(\cdot, 0, 1)$ ensures the values remain valid:
\begin{equation}
\tilde{\mathbf{S}} = \text{clip}(\mathbf{S} + \epsilon, 0, 1), \quad \epsilon \sim \mathcal{N}(0, \sigma^2\mathbf{I}).
\end{equation}

Given normalized image features $\mathbf{V}$ and text features $\mathbf{T}$, we compute the cosine similarity matrix $\mathbf{M} = \mathbf{T}\mathbf{V}^\top$. The text-to-image soft labels are derived as:
\begin{equation}
\mathbf{L}^{\text{t2i}} = \max(\mathbf{Y}, \tilde{\mathbf{S}}),
\end{equation}
where $\mathbf{Y}_{ij} = \mathbb{I}[y_i = y_j]$ is the identity matrix with diagonals set to 1.

The adaptive weights for text-to-image direction are computed as:
\begin{equation}
\alpha^{\text{t2i}} = \frac{\exp(\mathbf{M}/\tau) \odot \mathbf{L}^{\text{t2i}}}{\sum_j \exp(\mathbf{M}_{ij}/\tau) \mathbf{L}^{\text{t2i}}_{ij}}.
\end{equation}

The text-to-image loss is formulated as:
\begin{equation}
\mathcal{L}_{\text{t2i}} = \sum_i \left[ m -\langle \alpha^{\text{t2i}}_i, \mathbf{M}_i \rangle + \tau \log \sum_j \exp(\mathbf{M}_{ij}/\tau) (1 - \mathbf{L}^{\text{t2i}}_{ij}) \right]_{+},
\end{equation}
where $[\cdot]_+ = \max(0, \cdot)$. $\tau$ is a fixed temperature hyperparameter that controls the sharpness of the similarity distribution. A smaller $\tau$ sharpens the distribution and enforces harder discrimination among candidates~\cite{rde}. The image-to-text loss $\mathcal{L}_{\text{i2t}}$ is computed similarly using $\mathbf{M}^\top$ and $\mathbf{L}^{\text{i2t}} = \max(\mathbf{Y}^\top, \tilde{\mathbf{S}}^\top)$. The final soft TAL loss is $\mathcal{L}_{\text{stal}} = \mathcal{L}_{\text{t2i}} + \mathcal{L}_{\text{i2t}}$. Crucially, the margin parameter $m$ in the alignment loss is dynamically set according to the granularity level $g$.

\textbf{Discussion.} This module introduces a probabilistic reformulation of cross-modal matching that explicitly addresses the statistical nature of coarse-grained retrieval. The soft labeling mechanism models the posterior probability of semantic validity, where attribute overlap provides the likelihood term. The injected Gaussian noise accounts for epistemic uncertainty in real-world annotations. Mathematically, our \(\mathcal{L}_{\text{stal}}\) generalizes the standard triplet alignment loss~\cite{rde}, reducing to it when soft labels become binary, thus providing a principled framework for many-to-many matching. Additionally, the dynamic margin calibration ensures that the optimization intensity aligns with the discriminative demands at each granularity level.

\subsection{Granularity-Consistent Reasoning}
\label{sec6.4}

To further enhance the discriminative capability of visual features across granularities and enforce cross-modal granularity consistency, we introduce a Granularity-Consistent Collaborative Reasoning module comprising two complementary components.

\textit{Cross-modal Granularity Classification Head.} We design a shared granularity classifier that processes both visual and textual representations to predict their respective granularity levels. Formally, given a visual feature $\mathbf{v}_{cls}^g$ (extracted by activating expert $g$) and a text feature $\mathbf{t}_{cls}$, the granularity classification losses are defined as:
\begin{align}
    \mathcal{L}_{\text{g-cls}}^v &= \text{CE}(\text{Classifier}(\mathbf{v}_{cls}^g), g) \\
    \mathcal{L}_{\text{g-cls}}^t &= \text{CE}(\text{Classifier}(\mathbf{t}_{cls}), g),
\end{align}
where $\text{CE}$ denotes the cross-entropy loss and $\text{Classifier}$ is a single shared linear layer that projects the input feature into a $G$-dimensional logit space for granularity prediction. This explicit granularity prediction task enhances the discriminative power of both visual and textual features while promoting cross-modal alignment in the granularity space.

\textit{Granularity Consistency Matching Head.} We further introduce a binary matching head to verify granularity consistency between image-text pairs. Given an image-text pair $(I, T)$ with granularity label $g$, we construct positive and negative pairs as follows. 

Positive: $(\mathbf{v}_{cls}^g,\mathbf{t}_{cls})$ where the visual feature is extracted using the correct expert $g$. Negative: $(\mathbf{v}_{cls}^{g'}, \mathbf{t}_{cls})$ where $g' \neq g$ is randomly sampled from other granularity levels.

The granularity consistency matching loss is defined as:
\begin{equation}
\mathcal{L}_{\text{g-m}} = \text{BCE}(f_{\text{m}}(\mathbf{v}_{cls}^g, \mathbf{t}_{cls}), 1) + \text{BCE}(f_{\text{m}}(\mathbf{v}_{cls}^{g'}, \mathbf{t}_{cls}), 0),
\end{equation}
where $f_{\text{m}}$ is a bilinear matching function with sigmoid activation, $\text{BCE}$ is the binary cross-entropy loss, and $g' \neq g$ denotes a randomly sampled granularity level.

The complete granularity reasoning objective combines both components:
\begin{equation}
    \mathcal{L}_{\text{reason}} = \mathcal{L}_{\text{g-cls}}^v + \mathcal{L}_{\text{g-cls}}^t + \mathcal{L}_{\text{g-m}}.
\end{equation}

\textbf{Discussion.} This module introduces complementary inductive biases through multi-task learning. The shared granularity classifier enforces cross-modal alignment in a common granularity semantic space, narrowing the granularity distribution gap between vision and language. Concurrently, the binary matching head models the manifold of granularity-consistent image-text pairs. This dual-path optimization jointly constrains feature learning, enhancing discriminability for both identity matching and granularity prediction, thereby improving generalization across the granularity spectrum.

\subsection{Training and Inference}
\label{sec6.5}
The proposed CMAM framework is trained end-to-end with a unified objective function that integrates all proposed components. The overall training loss is formulated as:
\begin{equation}
\mathcal{L}_{\text{total}} = \mathcal{L}_{\text{stal}} + \lambda_1\mathcal{L}_{\text{ortho}} + \lambda_2\mathcal{L}_{\text{reason}},
\end{equation}
where \(\mathcal{L}_{\text{stal}}\) denotes the soft triplet alignment loss, \(\mathcal{L}_{\text{ortho}}\) the orthogonality constraint, \(\mathcal{L}_{\text{reason}}\) the granularity reasoning loss, and \(\lambda_1\), \(\lambda_2\) are balancing hyperparameters.

For datasets with multi-grained annotations like UFine6926-MG, we directly employ this training scheme. For conventional datasets~\cite{cuhkpedes,icfgpedes} lacking granularity diversity, we leverage our proposed annotation engine to automatically generate multi-grained texts, enabling consistent training methodology across different data sources.

During inference, we adopt a multi-gallery retrieval strategy. For each image in the gallery, we generate multi-grained visual representations by sequentially activating all experts in the visual encoder:
\begin{equation}
\mathbf{v}_{cls}^g = \text{VE}(I;g), \quad \forall g \in \{1, 2, \dots, G\}.
\end{equation}
This creates \(G\) parallel galleries \(\{\mathcal{G}_1, \mathcal{G}_2, \dots, \mathcal{G}_G\}\), each corresponding to a specific granularity level.

Given a text query, the retrieval process proceeds as follows. For benchmarks with granularity labels (e.g., MG-Eval), we directly use the provided label \(g\) to select the corresponding gallery \(\mathcal{G}_g\). For conventional datasets without granularity annotations, we predict the granularity level using our trained classifier: \(\hat{g} = \arg\max \text{Classifier}(\mathbf{t_{cls}})\), then retrieve from gallery \(\mathcal{G}_{\hat{g}}\).

The image-text similarity is computed as the cosine similarity between their global representations:
\begin{equation}
S(I,T) = \frac{\mathbf{v}_{cls}^g \cdot \mathbf{t}_{cls}}{\|\mathbf{v}_{cls}^g\| \|\mathbf{t}_{cls}\|},
\end{equation}
where \(\mathbf{v}_{cls}^g\) is the visual feature from the selected gallery and \(\mathbf{t}_{cls}\) is the global text feature. This multi-gallery inference paradigm ensures that each query is matched against visually compatible representations at the appropriate specificity level, providing a principled solution for handling the granularity spectrum in real-world retrieval scenarios.

\section{Experiments}
\subsection{Datasets and Evaluation Metrics}
We augment existing datasets with multi-grained text annotations using our proposed engine, resulting in CUHK-PEDES-MG~\cite{cuhkpedes}, ICFG-PEDES-MG~\cite{icfgpedes}, and our extended UFine6926-MG. These multi-grained texts are applied exclusively to training sets, while preserving original test sets unmodified for fair comparison with prior works. Beyond the standard evaluation paradigm, we introduce two additional protocols: 1) Fixed-granularity Evaluation: Models trained on multi-grained data are evaluated on original test sets using standard metrics (R@1, R@5, R@10, mAP). 2) Multi-Granularity Evaluation: Comprehensive assessment on our MG-Eval using Separate/Progressive Granularity Assessments, reporting both standard metrics and our proposed mSD. This dual-protocol design ensures direct comparability with existing methods while thoroughly evaluating any-granularity retrieval capability.

\begin{table*}[t]\small
    \centering
    \caption{Performance comparison on standard single-grained benchmarks. The upper block lists methods trained on the original fixed-granularity data, while the lower block employs our augmented multi-grained data. For fair comparison, methods~\cite{tan2024harnessing,jiang2025modeling,qin2025human} leveraging MLLMs and generation techniques for large-scale training data expansion are excluded from this table.}
    \small{
    \resizebox{0.98\linewidth}{!}{
        \renewcommand\arraystretch{1.2}

    \begin{tabular}{llcccccccccccc}
    \hline\thickhline
    &  & \multicolumn{4}{c}{\textbf{CUHK-PEDES}} & \multicolumn{4}{c}{\textbf{ICFG-PEDES}} & \multicolumn{4}{c}{\textbf{UFine6926}} \\  
    \multirow{-2}{*}{\textbf{Methods}}    &   \multirow{-2}{*}{\textbf{Reference}}       & R@1   & R@5   & R@10   & mAP    & R@1   & R@5   & R@10   & mAP    & R@1   & R@5   & R@10  & mAP   \\ 
    \hline
    \multicolumn{14}{l}{\textit{Training with Original Fixed-Granularity Data}} \\
    SSAN~\cite{icfgpedes} & Arxiv21 & 61.37 &80.15 &86.73&-&54.23 &72.63 &79.53&-&75.09 &88.63 &92.84& 73.14\\
    LGUR~\cite{shao2022learning} & MM22 & 65.25 &83.12 &89.00 &- &59.02& 75.32& 81.56&-&70.69& 84.57& 89.91& 68.93 \\
    CFine~\cite{yan2023clip}   & TIP23                      & 69.57 & 85.93 & 91.15 & -      & 60.83 & 76.55 & 82.42 & -      & -&-&-&-     \\
    IRRA~\cite{irra} & CVPR23                     & 73.38 & 89.93 & 93.71 & 66.13  & 63.46 & 80.25 & 85.82 & 38.06  & 85.02& 94.31& 96.75& 83.91  \\
    BiLMa~\cite{fujii2023bilma} & ICCV23                    & 74.03 & 89.59 & 93.62 & 66.57  & 63.83 & 80.15 & 85.74 & 38.26  & -&-&-&-  \\
    RaSa~\cite{bai2023rasa}    & IJCAI23                    & 76.51 & 90.29 & 94.25 & 69.38  & 65.28 & 80.40 & 85.12 & 41.29  & -&-&-&-  \\
    TBPS~\cite{cao2024empirical}& AAAI24               & 73.54 & 88.19 & 92.35 & 65.38  & 65.05 & 80.34 & 85.47 & 39.83  &  84.27&93.28&96.60&84.22  \\ 
    RDE~\cite{rde}    & CVPR24                     & 75.94 & 90.14 & 94.12 & 67.56  & 67.68 & 82.47 & 87.36 & 40.06  & 87.60 &95.65 &97.46 &86.10 \\ 
    HKGR~\cite{zeng2025hierarchical} & NN25&  75.21 &90.22 & 94.31 &67.28 &65.29& 81.17& 86.22 &39.40 & -&-&-&-\\
    SDG~\cite{gou2025instance} &KBS25&74.68& 89.39& 93.57& 68.10&66.56& 81.40& 86.51& 43.43& -&-&-&-\\
    FGCLIP~\cite{xie2025fgclip}& ICML25    & 74.69 & 89.49 & 93.32 & 66.82   & 65.32 & 81.22 & 86.65 & 39.47  & 85.80   &94.55    &96.85    & 83.86     \\
    \hdashline
    CFAM~\cite{ufinebench}& CVPR24                   & 75.60 & 90.53 & 94.36 & 67.27  & 65.38 & 81.17 & 86.35 & 39.42  & 85.55& 94.51& 97.02& 84.23   \\
    \rowcolor{lightgray!50}
    \textbf{CMAM}& \textbf{Ours}             & \textbf{76.62} & \textbf{91.27} & \textbf{94.33}  & \textbf{69.42}   & \textbf{67.91} & \textbf{82.66} &\textbf{87.42}  & \textbf{41.63}  & \textbf{88.21}   &\textbf{95.87}    &\textbf{97.84}    & \textbf{87.16}     \\
    \hline
    \multicolumn{14}{l}{\textit{Training with Augmented Multi-Granularity Data}} \\
    CLIP~\cite{clip}& ICML21                    & 59.26 & 79.32 & 85.95  & 54.00   & 53.67 & 72.75 &79.56  & 29.88  & 78.33   &90.65    &94.31    & 76.01     \\
    IRRA~\cite{irra}& CVPR23                    & 67.12 & 85.12 & 90.76  & 61.86   & 59.14 & 76.71 &83.04  & 37.27  & 81.75   &92.71    &95.69    & 80.58     \\
    TBPS~\cite{cao2024empirical} & AAAI24                    & 66.21 & 83.76 & 88.98  & 60.13   & 57.47 & 74.28 &81.47  & 32.36  & 79.66   &90.58    &94.27    & 75.25     \\
    RDE~\cite{rde} & CVPR24                    & 69.31 & 86.21 & 90.79  & 61.73   & 59.19 & 77.17 &83.22  & 33.02  & 79.26   &91.29    &94.83    & 75.78     \\
    FGCLIP~\cite{xie2025fgclip}& ICML25                    & 69.57  &  86.11  &  91.13  &  61.69   & 60.10 & 77.51 & 83.58 & 34.02  & 78.24   &91.00    &94.56    & 74.97     \\
    \hdashline
    CFAM~\cite{ufinebench}& CVPR24                    & 68.41 & 85.21 & 90.38  & 61.46   & 59.32& 77.26 &82.96  & 33.54  & 82.06   &91.86    &95.13    & 76.65     \\
    \rowcolor{lightgray!50}
    \textbf{CMAM}& \textbf{Ours}             & \textbf{77.12} & \textbf{92.03} & \textbf{94.84}  & \textbf{69.88}   & \textbf{68.13} & \textbf{82.96} &\textbf{87.67}  & \textbf{41.92}  & \textbf{88.45}   &\textbf{96.12}    &\textbf{98.02}    & \textbf{87.73}     \\
    \hline\thickhline
    \end{tabular}
    }}
     \label{tab:sota tipr}
\end{table*}

\subsection{Implementation Details}
In our implementation, we adopt the pre-trained cross-modal ViT-Base~\cite{xie2025fgclip} as the backbone for our vision-language encoder to leverage its robust representative capabilities. To handle multi-grained alignment, five heterogeneous experts are instantiated with dimensionalities of 192, 384, 768, 1536, and 3072. The balancing coefficients $\lambda_1$ and $\lambda_2$ in the loss function are both empirically set to 1. The framework is optimized using the Adam optimizer with a total of 30 epochs, including a 2-epoch linear warm-up phase. We adopt a cosine learning rate decay strategy, where the learning rate is decayed to 10\% of its initial value at the 10th and 25th epochs. The initial learning rate for pre-trained modules is set to $1 \times 10^{-5}$, while the experts and the granularity-consistent reasoning module are assigned a higher learning rate of $2 \times 10^{-5}$. The training batch size is 128. Input images are resized to $384 \times 128$. For visual data, we utilize random horizontal flipping, random cropping with padding, and random erasing. For textual data, we apply token-level augmentations, including random masking, replacement, and removal. In the PCIA, the noise intensity $\sigma$, the threshold $\theta$ and the temperature $\tau$ are set to 0.1, 0.6 and 0.015, respectively, to model the inherent semantic ambiguity.

\vspace{-2mm}
\subsection{Comparison with State-of-the-Art Methods}
This section compares the proposed CMAM with state-of-the-art methods on both single-grained and multi-grained text-based person retrieval. As the first method designed for multi-grained TPR, CMAM is first evaluated against single-grained methods on conventional benchmarks. We then extend the comparison to our proposed MG-Eval benchmark to assess multi-grained retrieval performance.

\noindent
\textbf{Performance Comparison on Single-grained TPR.} As reported in Tab.~\ref{tab:sota tipr}, we conduct a comprehensive evaluation on standard single-grained benchmarks under two distinct training paradigms: using the original fixed-granularity data and our augmented multi-grained data. To ensure a rigorous comparison, CMAM is evaluated under both settings. A critical observation highlights the divergent adaptability of different frameworks: while strong baselines (e.g., IRRA~\cite{irra} and RDE~\cite{rde}) exhibit notable performance degradation when trained on multi-grained data, succumbing to the increased semantic variance, CMAM conversely achieves a slight performance gain. This phenomenon indicates that existing methods struggle to disentangle valid multi-granular signals from noise, whereas CMAM effectively leverages granularity diversity for representation enhancement. Ultimately, CMAM secures state-of-the-art or highly competitive performance across all datasets (e.g., achieving 77.12\% rank-1 on CUHK-PEDES), confirming its superiority in granularity-complex scenarios.

\begin{table}[t]
    \centering
    \caption{Performance comparison on the multi-grained MG-Eval benchmark using the Separate Granularity Assessment. All models are trained on the UFine6926-MG dataset.}
    {
    \resizebox{\linewidth}{!}{
        \renewcommand\arraystretch{1.2}
    
    \begin{tabular}{llccccc}
    \hline\thickhline
    &  & \multicolumn{5}{c}{\textbf{MG-Eval (Separate)}} \\ 
    \multirow{-2}{*}{\textbf{Methods}}    &   \multirow{-2}{*}{\textbf{Reference}}       & R@1   & R@5   & R@10   & mAP  & mSD    \\
    \hline
    CLIP~\cite{clip}   & ICML21    & 36.90  & 55.06  & 62.78 &20.79 & 15.97 \\
    IRRA~\cite{irra}& CVPR23   & 42.10  & 60.92  & 69.02 &25.31 & 19.45 \\
    TBPS~\cite{cao2024empirical} & AAAI24  & 40.48  & 61.24  & 68.43 &24.31 & 16.42 \\
    RDE~\cite{rde}   & CVPR24     & 42.78  & 62.86  & 71.16 &22.62 & 16.68 \\
    FGCLIP~\cite{xie2025fgclip} & ICML25   & 45.12  & 64.64  & 72.30 &23.68 & 17.14 \\
    \hdashline
    CFAM~\cite{ufinebench} & CVPR24   & 43.12  & 62.38  & 70.46 &24.67 & 17.29 \\
    \rowcolor{lightgray!50}
    \textbf{CMAM}                         & \textbf{Ours}         & \textbf{50.40}  & \textbf{67.70}  & \textbf{75.02} &\textbf{28.57} & \textbf{21.46} \\ 

    \hline\thickhline
    \end{tabular}
    }}
    \label{tab:MGEVal-S}
    \vspace{-3mm}
\end{table}

\begin{figure*}[htb]
\centering
\includegraphics[width=\linewidth]{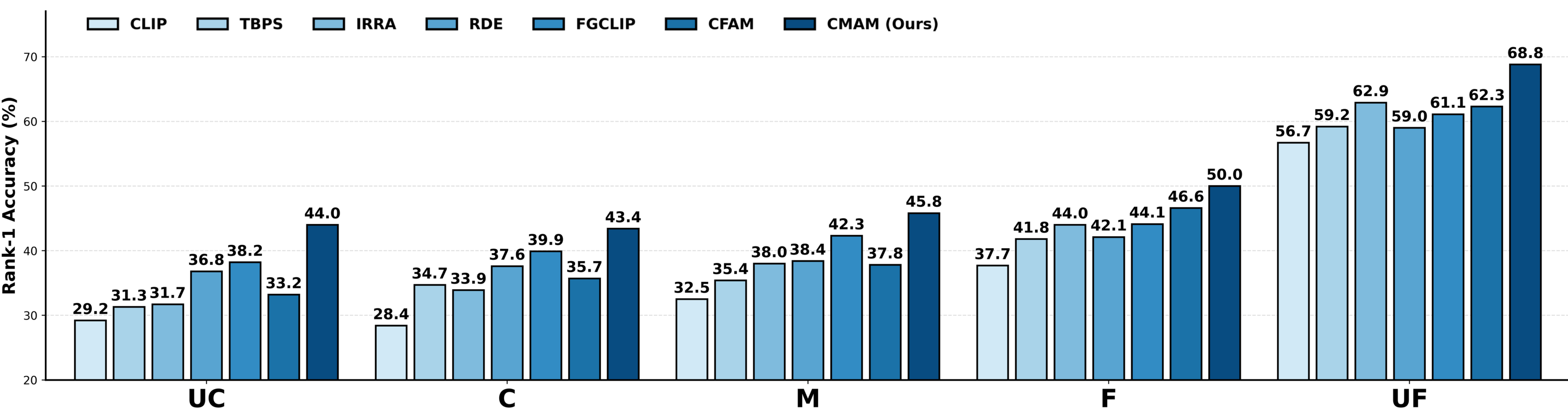}
\caption{Performance comparison across five granularity levels. We report the rank-1 accuracy on the MG-Eval benchmark, spanning from ultra coarse to ultra fine. Our CMAM achieves best results across the entire spectrum, outperforming existing methods in both information sparse and rich retrieval scenarios, validating its effectiveness for any-granularity person retrieval.}
\label{fig:AGR}
\vspace{-2mm}
\end{figure*}

\noindent
\textbf{Performance Comparison on Multi-grained TPR.} We further evaluate the model's capability under the more challenging multi-grained retrieval scenario. All methods are trained on our UFine6926-MG dataset and tested on the proposed MG-Eval benchmark under two protocols. As shown in Tab.~\ref{tab:MGEVal-S}, CMAM achieves state-of-the-art performance in the separate assessment, outperforming all reproduced strong baselines. Notably, our CMAM attains a rank-1 accuracy of 50.40\%, outperforming the strongest baseline, FGCLIP~\cite{xie2025fgclip}, by 5.28\%. Also, the significant improvement in mSD is particularly noteworthy, as it indicates that CMAM not only identifies the correct targets but also produces a more discriminative and calibrated similarity distribution. The progressive assessment results are presented in Tab.~\ref{tab:MGEval-P}, where models are assessed under different reduction rates of 12.5\%, 25\%, and 50\%. A lower rate imposes higher demands on the model's discriminative ability. CMAM consistently ranks top under all settings, showing particular strength in the most challenging 12.5\% reduction scenario. These results collectively confirm that our CMAM not only excels in fine-grained matching but also robustly handles the dynamic nature of multi-grained textual queries.

\begin{figure}[htb]
\centering
\vspace{-2mm}
\includegraphics[width=\linewidth]{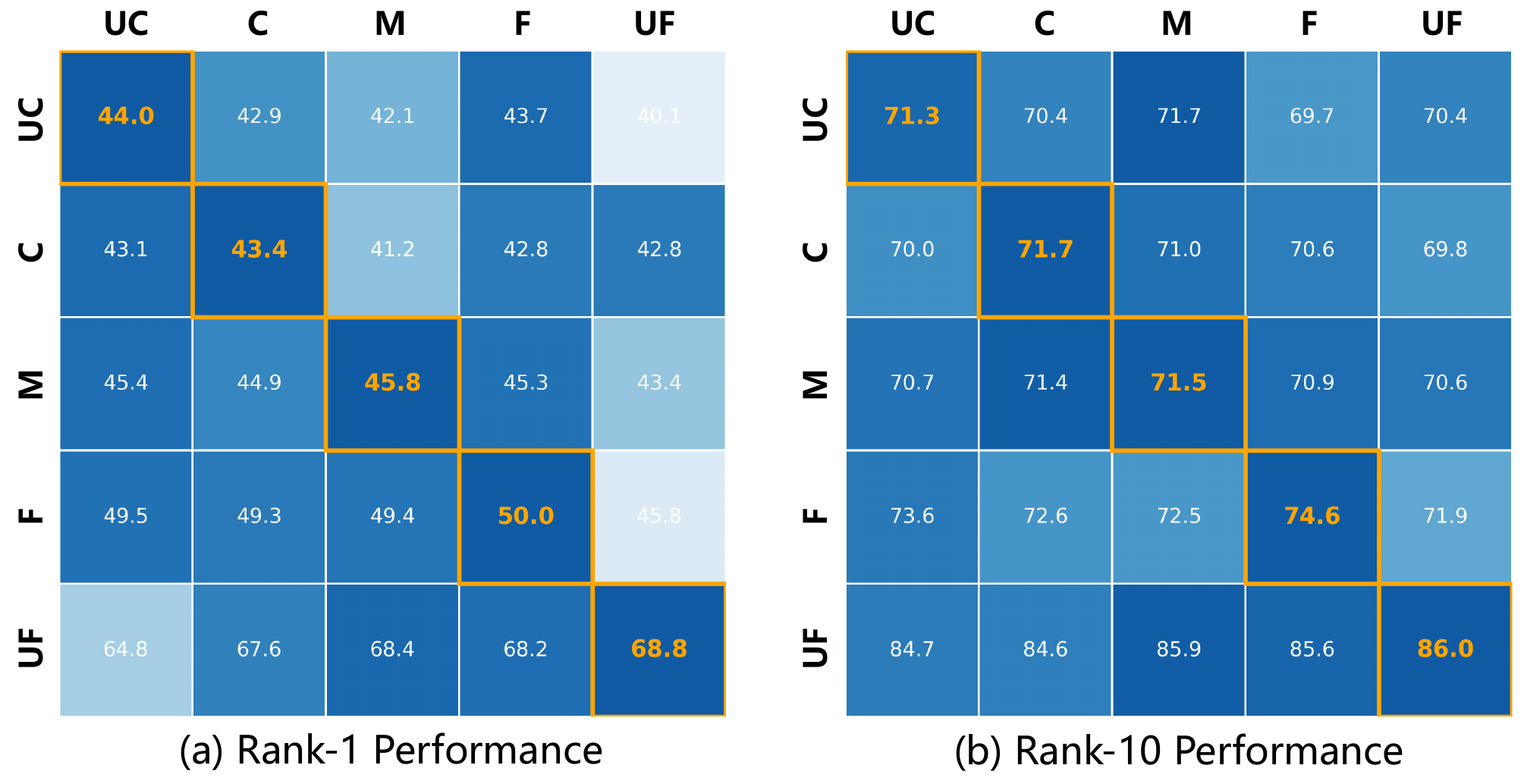}
\caption{Robustness analysis across granularities. Heatmaps of (a) Rank-1 and (b) Rank-10 accuracy across all combinations of textual granularities (rows) and visual experts (columns). }
\label{fig:consistent}
\vspace{-3mm}
\end{figure}

\subsection{Analysis of Any Granularity Retrieval}
\begin{table}[t]
    \centering
    \caption{Performance comparison on the multi-grained MG-Eval benchmark using the Progressive Granularity Assessment. All models are trained on the UFine6926-MG dataset.}
    \large{
    \resizebox{\linewidth}{!}{
        \renewcommand\arraystretch{1.2}
    \begin{tabular}{cllccccc}
    \hline\thickhline
   & &  & \multicolumn{5}{c}{\textbf{MG-Eval (Progressive)}} \\ 
    \multirow{-2}{*}{\textbf{Rate $\rho$}}&\multirow{-2}{*}{\textbf{Methods}}    &   \multirow{-2}{*}{\textbf{Reference}}       & R@1   & R@5   & R@10   & mAP  & mSD   \\
    \hline
    \multirow{7}{*}{12.5\%} 
        &CLIP~\cite{clip}   & ICML21    & 55.20  & 70.50  & 74.90 &22.97 & 16.76 \\
    &IRRA~\cite{irra}& CVPR23   & 61.10  & 75.00  & 78.50 &27.41 & 20.17 \\
    &TBPS~\cite{cao2024empirical}     & AAAI24  & 59.60  & 73.20  & 77.10 &23.58 & 17.26 \\
    &RDE~\cite{rde}   & CVPR24     & 63.60  & 77.50  & 81.30 &25.05 & 18.36 \\
    &FGCLIP~\cite{xie2025fgclip} & ICML25   & 63.60  & 78.90  & 82.20 &26.30 & 19.31 \\
        \cdashline{2-8}
        & CFAM~\cite{ufinebench}       & CVPR24    & 61.70  & 75.80  & 79.40 &25.12 & 18.68 \\
        \rowcolor{lightgray!50}
        \cellcolor[HTML]{FFFFFF}
        & \textbf{CMAM}  &   \textbf{Ours}      & \textbf{70.10} & \textbf{80.80} & \textbf{83.60} & \textbf{30.44} & \textbf{22.51} \\
        \hline
        
    \multirow{7}{*}{25\%} 
        &CLIP~\cite{clip}   & ICML21    & 62.70  & 81.80  & 85.70 &27.67 & 20.87 \\
    &IRRA~\cite{irra}& CVPR23   & 70.50  & 85.10  & 89.10 &32.21 & 24.47 \\
    &TBPS~\cite{cao2024empirical}     & AAAI24  & 66.30  & 83.20  & 88.40 &28.26 & 21.57 \\
    &RDE~\cite{rde}   & CVPR24     & 68.20  & 86.00  & 90.00 &29.16 & 22.03 \\
    &FGCLIP~\cite{xie2025fgclip} & ICML25   & 69.90  & 85.90  & 89.80 &30.31 & 22.93 \\
        \cdashline{2-8}
        & CFAM~\cite{ufinebench}       & CVPR24    & 69.20  & 85.60  & 88.80 &29.37 & 22.68 \\
        \rowcolor{lightgray!50}
        \cellcolor[HTML]{FFFFFF}
        &\textbf{CMAM}  &   \textbf{Ours}      & \textbf{75.90} & \textbf{88.30} & \textbf{92.20} & \textbf{35.59} & \textbf{27.12} \\
        \hline
        
    \multirow{7}{*}{50\%} 
        &CLIP~\cite{clip}   & ICML21    & 64.10  & 82.30  & 87.10 &29.93 & 22.52 \\
    &IRRA~\cite{irra}& CVPR23   & 71.50  & 86.10  & 91.40 &34.33 & 25.97 \\
    &TBPS~\cite{cao2024empirical}     & AAAI24  & 67.10  & 83.40  & 89.60 &30.12 & 22.96 \\
    &RDE~\cite{rde}   & CVPR24     & 68.10  & 86.50  & 90.90 &30.71 & 23.10 \\
    &FGCLIP~\cite{xie2025fgclip} & ICML25   & 68.20  & 85.00  & 89.90 &30.76 & 23.14 \\
        \cdashline{2-8}
        & CFAM~\cite{ufinebench}       & CVPR24    & 68.80  & 85.70  & 90.20 &30.28 & 23.04 \\
        \rowcolor{lightgray!50}
        \cellcolor[HTML]{FFFFFF}
        & \textbf{CMAM}  &   \textbf{Ours}      & \textbf{75.90} & \textbf{89.70} & \textbf{92.60 }& \textbf{37.12} & \textbf{28.07} \\
    
    \hline\thickhline
    \end{tabular}
    }}
    \label{tab:MGEval-P}
    \vspace{-2mm}
\end{table}

\textbf{Adaptability to Any Granularity.} We compare CMAM with six SOTA methods across the MG-Eval spectrum. As shown in Fig.~\ref{fig:AGR}, all methods exhibit a consistent upward trend in accuracy as query granularity increases from coarse to fine, confirming that richer descriptions provide more discriminative cues for matching. Crucially, CMAM maintains a significant lead over all competitors at every granularity level. While existing models often show performance brittleness when query informativeness shifts, CMAM demonstrates superior robustness and stability throughout. These results underscore its exceptional capability to handle the entire granularity spectrum, offering a versatile solution for real-world multi-grained retrieval.

\noindent
\textbf{Cross-Granularity Generalization and Robustness}. We evaluate CMAM's resilience to granularity misalignment by cross-pairing textual queries with visual experts across all five levels. As illustrated in Fig.~\ref{fig:consistent}, while optimal performance is achieved upon granularity alignment, CMAM exhibits remarkable stability across cross-granularity configurations, with minimal performance fluctuations. This suggests that the OE-MVP module successfully learns a granularity-resilient identity representation. Specifically, the granularity-shared expert preserves fundamental discriminative cues, ensuring that identity-level semantics remain invariant even when a sub-optimal granularity expert is engaged. Such robustness is crucial for real-world deployment, as it mitigates the risk of performance degradation stemming from imprecise query specificity estimation.

\begin{table*}[t]
\centering
\caption{Ablation study on the effectiveness of individual CMAM components. The baseline (\#0) is a standard dual-branch cross-modal framework without specialized designs, employing only the conventional TAL loss for image-text alignment.}
\label{tab:component_ablation}
\small{
\resizebox{0.95\linewidth}{!}{
\renewcommand\arraystretch{1.2}
\begin{tabular}{c|ccc|ccccc|ccccc}
\hline\thickhline 
\multicolumn{1}{c|}{\multirow{2}{*}{\textbf{\#}}} & \multicolumn{3}{c|}{\textbf{Components}}&\multicolumn{5}{c|}{\textbf{MG-Eval (Separate)}}&\multicolumn{5}{c}{\textbf{MG-Eval (Progressive, $\rho=50\%$)}}\\

& OE-MVP &PCIA &GCR&R@1&R@5&R@10&mAP&mSD&R@1&R@5&R@10&mAP&mSD    \\
\hline 
0 && &&44.98	&64.32&71.56	&23.07&16.83	&67.70&85.40&90.70&30.08&22.60 \\
1 &\checkmark& &&48.48	&66.40&74.14	&26.75&19.90	&75.60&88.60&92.20&35.47&26.76 \\
2 &\checkmark& \checkmark& &50.10	&67.50&74.44	&28.44&21.34	&75.80&89.60&92.60&36.95&27.94 \\
3&\checkmark& &\checkmark&49.14	&66.88&74.74	&28.37&21.21	&75.10&88.80&92.50&35.98&27.21 \\
\rowcolor{gray!20}
4&\checkmark&\checkmark&\checkmark&\textbf{50.40}  & \textbf{67.70}  & \textbf{75.02} &\textbf{28.57} & \textbf{21.46}	&\textbf{75.90} & \textbf{89.70} & \textbf{92.60 }& \textbf{37.12} & \textbf{28.07} \\
\hline\thickhline
\end{tabular}}}
\end{table*}

\begin{figure*}[htb]
\centering
\vspace{-1mm}
\includegraphics[width=\linewidth]{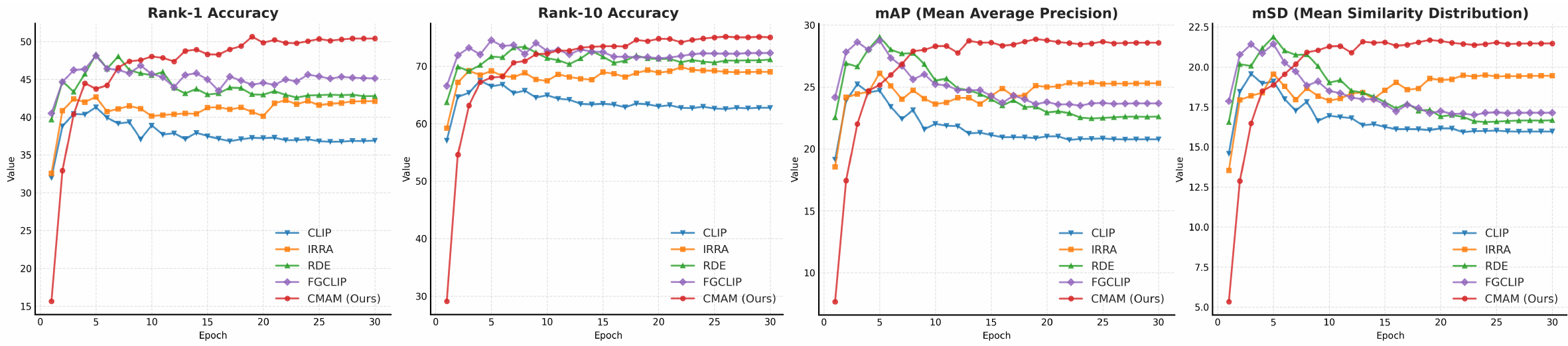}
\vspace{-6mm}
\caption{Comparison of training dynamics and convergence characteristics. While existing state-of-the-art methods~\cite{clip,irra,rde,xie2025fgclip} often exhibit performance fluctuations or late-stage degradation due to granularity-induced optimization conflicts, our proposed CMAM framework demonstrates stable, monotonic improvement and converges to a superior performance plateau.}
\vspace{-3mm}
\label{fig:training_curve}
\end{figure*}

\noindent
\textbf{Training Stability and Convergence.} Fig.~\ref{fig:training_curve} illustrates the training dynamics of CMAM compared with four representative methods on MG-Eval. Existing models generally follow an ``increase-then-decline'' pattern because they often struggle with optimization conflicts or semantic overfitting across different granularities. Conversely, CMAM demonstrates consistent monotonic improvement and converges to a stable high-performance plateau. This behavior confirms that our framework effectively harmonizes multi-grained signals while avoiding the performance degradation common in traditional approaches. Such reliable convergence highlights the robustness of CMAM and ensures predictable performance for practical deployment.

\vspace{-2mm}
\subsection{Ablation Studies}
\textbf{Effectiveness of Core Components.} We systematically evaluate the contributions of our designs via incremental ablation on the MG-Eval benchmark (Tab.~\ref{tab:component_ablation}). Starting from a baseline (\#0) utilizing a standard dual-branch architecture with TAL loss~\cite{rde}, the sequential integration of our proposed modules consistently improves retrieval performance. Most notably, the OE-MVP module emerges as the dominant contributor, yielding the most significant performance leap with a 3.68\% mAP improvement in the separate granularity assessment. This substantial gain validates the critical role of orthogonal experts in disentangling multi-grained visual features. Subsequently, the inclusion of PCIA and GCR further refines the alignment, with the full CMAM framework achieving superior performance. This trajectory underscores the complementary nature of these components, where OE-MVP establishes a robust foundation that is effectively polished by the other two modules.

\begin{table}[htb]
\centering
\caption{Ablation study on the performance contribution of individual components within the OE-MVP module.}
\label{tab:oemvp}
{
\resizebox{\linewidth}{!}{
\renewcommand\arraystretch{1.2}
\begin{tabular}{c|ccc|ccccc}
\hline\thickhline 
\multicolumn{1}{c|}{\multirow{2}{*}{\textbf{\#}}} & \multicolumn{3}{c|}{\textbf{OE-MVP}}&\multicolumn{5}{c}{\textbf{MG-Eval (Separate)}}\\
& MoE &Ortho.&Het.&R@1&R@5&R@10&mAP&mSD   \\
\hline 
0 && &&44.98	&64.32&71.56	&23.07&16.83	 \\
1 &\checkmark& &&45.36	&64.58&71.48	&23.12&16.96	 \\
2 &\checkmark& \checkmark& &47.60	&66.06&73.28	&25.98&19.21	 \\
3&\checkmark& &\checkmark&46.68	&65.42&72.84	&24.26&18.45	 \\
\rowcolor{gray!20}
4&\checkmark&\checkmark&\checkmark&\textbf{48.48}	&\textbf{66.40}&\textbf{74.14}	&\textbf{26.75}&\textbf{19.90} \\
\hline\thickhline
\end{tabular}}}
\end{table}

\begin{table}[htb]
    \centering
    \caption{Rank-1 accuracy (\%) on MG-Eval across granularity levels under varying attribute overlap threshold $\theta$ in PCIA.}
    \vspace{-1mm}
    \label{tab:theta}
    {
    \resizebox{\linewidth}{!}{
    \renewcommand\arraystretch{1.2}
    
    \begin{tabular}{lcccccc}
    \hline\thickhline
    &  \multicolumn{6}{c}{\textbf{Attribute Overlap Threshold $\theta$}} \\ 
    \multirow{-2}{*}{\textbf{Granularity}} & 0.3   & 0.4   & 0.5   & 0.6  & 0.7 & 0.8  \\
    \hline
    Ultra Coarse & 41.50  & 42.60  & 43.10 &43.10 & 42.60&42.30 \\
    Coarse & 41.80  & 42.40  & 43.00 &43.20 & 42.90&42.50 \\
    Medium  & 45.20  &45.90  & 46.00 &46.10 & 46.00&45.80 \\
    Fine  & 46.30  & 46.80  & 46.90 &47.00 & 47.10&46.90 \\
    Ultra Fine & 66.80  & 67.70  & 67.90 &68.10 & 68.00&68.00 \\
    \hline
    Average & 48.32  & 49.08  & 49.38 &49.50 & 49.32&49.10 \\

    \hline\thickhline
    \end{tabular}
    }}
    \vspace{-5mm}
\end{table}

\noindent
\textbf{Ablation on OE-MVP Design Choices.} We evaluate the architectural components of OE-MVP on the MG-Eval benchmark with PCIA and GCR deactivated for a controlled assessment. As detailed in Tab.~\ref{tab:oemvp}, a naive expansion of model capacity using a basic MoE with identical experts (\#1) yields negligible performance gains compared to the baseline (\#0). This stagnation suggests that without explicit regularization, multiple experts tend to degenerate into learning redundant representations. In sharp contrast, introducing the orthogonality loss (\#2) triggers a substantial performance leap, boosting mAP by 2.72\%. This result identifies the orthogonality constraint as the pivotal mechanism that enforces functional specialization, effectively converting increased capacity into discriminative power. Ultimately, the full design combining structural heterogeneity and orthogonality (\#4) achieves the highest accuracy.

\begin{figure*}[t]
\centering
\includegraphics[width=\linewidth]{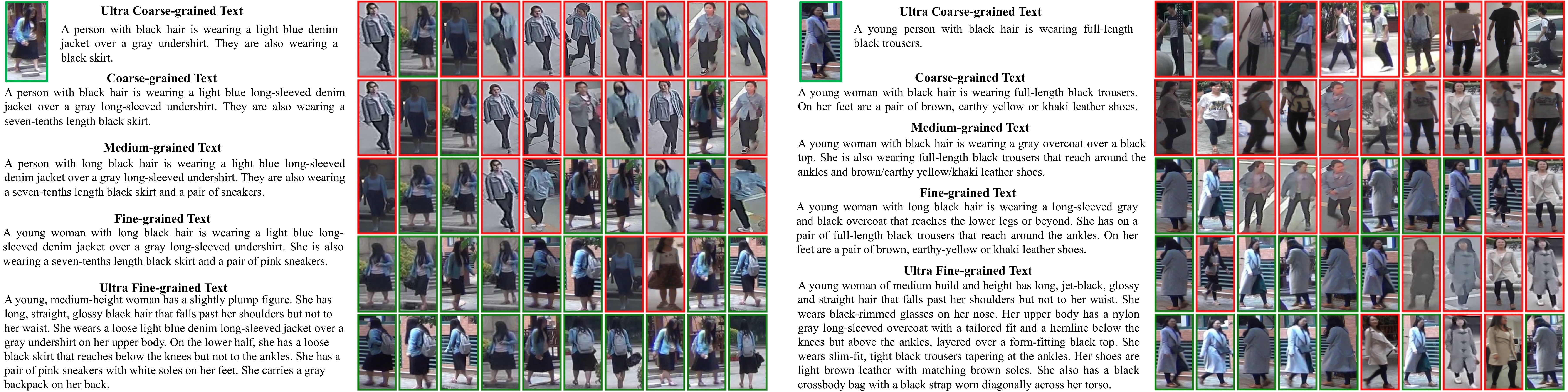}
\vspace{-5mm}
\caption{Qualitative visualization of the Top-10 retrieval results across the granularity spectrum. For each representative identity, the retrieved gallery images progressively converge toward the true target as the textual query shifts from ultra coarse to ultra fine. In coarse-grained scenarios, CMAM retrieves multiple visually similar candidates that match the sparse attributes; in fine-grained scenarios, the model effectively leverages detailed cues to precisely isolate the ground-truth identity.}
\vspace{-4mm}
\label{fig:example}
\end{figure*}

\begin{figure}[htb]
\centering
\vspace{-2mm}
\includegraphics[width=0.94\linewidth]{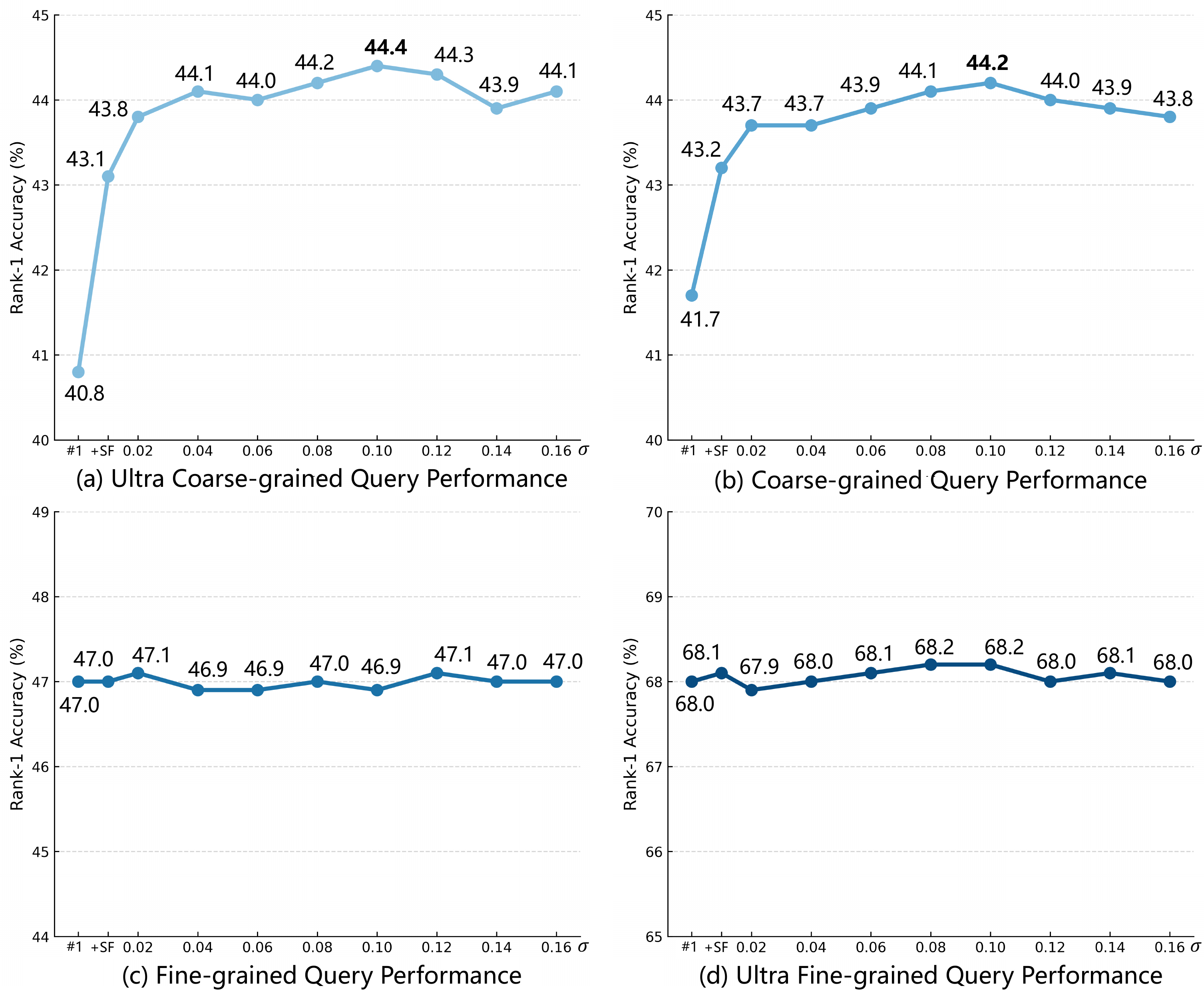}
\vspace{-1mm}
\caption{Impact of soft label and noise scale $\sigma$ in PCIA across various granularities. PCIA selectively improves performance on ultra-coarse and coarse queries by addressing one-to-many ambiguity, without sacrificing precision on finer descriptions.}
\label{fig:pcia}
\vspace{-8mm}
\end{figure}

\noindent
\textbf{Ablation and Sensitivity Analysis on PCIA.} 
Using the OE-MVP-equipped model with hard labels as baseline
(\#1 in Tab.~\ref{tab:component_ablation}), we conduct a
two-stage sensitivity analysis to determine the optimal
configuration of PCIA.

\noindent\textit{a) Sensitivity to $\theta$.}
We vary the attribute overlap threshold $\theta \in \{0.3,
0.4, 0.5, 0.6, 0.7, 0.8\}$. As reported in
Tab.~\ref{tab:theta}, coarse and ultra-coarse queries are most
sensitive to $\theta$: too small a value (\emph{e.g.},
$\theta{=}0.3$) introduces semantically irrelevant
cross-identity labels as noise, while too large a value
(\emph{e.g.}, $\theta{\geq}0.7$) suppresses valid
cross-identity correspondences, causing PCIA to degenerate
toward hard label supervision. Fine-grained queries remain
robust across all settings due to their naturally unambiguous
one-to-one correspondences. Performance peaks at
$\theta{=}0.6$, which we fix as the default soft label
configuration (denoted ``+SF'').

\noindent\textit{b) Sensitivity to $\sigma$.}
Building on ``+SF'', we examine Gaussian noise injection by
varying $\sigma$. As shown in Fig.~\ref{fig:pcia}, ``+SF'' alone
already yields consistent gains on coarse queries, confirming
that soft labels effectively resolve one-to-many matching
ambiguity. Additional noise injection peaks at $\sigma{=}0.1$:
smaller values fail to capture real-world annotation
uncertainty, while larger values corrupt the soft label
structure. Fine and ultra-fine query performance remains
stable throughout, demonstrating that PCIA's probabilistic
regularization is targeted and does not compromise
high-granularity retrieval precision.

\begin{figure}[t]
\centering
\includegraphics[width=0.99\linewidth]{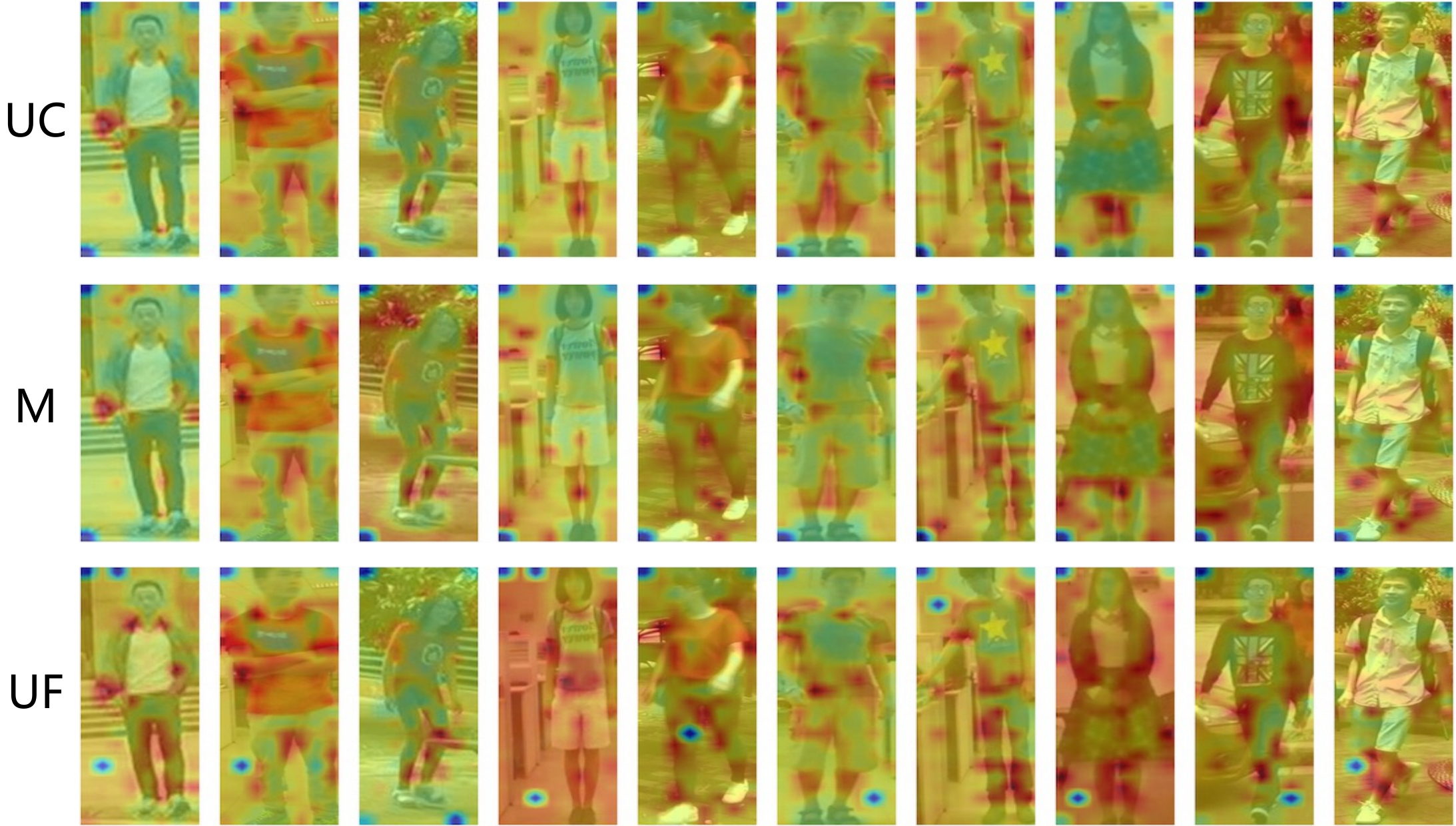}
\caption{Perceptual evolution across granularities. Feature activation maps show that fine-grained experts exhibit intensified and localized responses to subtle attributes, whereas coarse-grained experts yield relatively sparse and subdued activations.}
\vspace{-5mm}
\label{fig:attn}
\end{figure}
\subsection{Qualitative Analysis}

\textbf{Qualitative Analysis of Multi-granularity Retrieval.} To intuitively evaluate the retrieval behavior of CMAM, we visualize the Top-10 results for two representative identities across the five-level granularity spectrum in Fig.~\ref{fig:example}. As the query progresses from ultra-coarse to ultra-fine, CMAM exhibits a clear progressive refinement in retrieval precision. At the coarsest levels, while the Top-10 results may include different individuals, they consistently share high semantic similarity with the target’s general attributes, demonstrating the model's capacity to navigate identity-level ambiguity. As specific descriptive details are introduced, the candidate pool monotonically converges toward the unique ground-truth target. This qualitative trend confirms that CMAM effectively harnesses the informativeness gradient within the textual descriptions, successfully bridging the semantic gap regardless of query specificity.

\noindent
\textbf{Visualization of Multi-granularity Feature Activation.} To examine the expert specialization, we visualize the $L_2$ normalization of hidden states from the final layer, which serves as a proxy for the model's perceptual saliency (Fig.~\ref{fig:attn}). We observe a clear progression in both spatial coverage and activation intensity: coarse-grained experts exhibit relatively sparse and subdued responses, focusing only on fragmentary body regions with lower magnitude. In contrast, fine-grained experts demonstrate a significantly sharpened focus, characterized by high-intensity activations that precisely localize on subtle, discriminative attributes. This phenomenon indicates that our orthogonal constraints effectively guide the experts to adapt their feature-extracting effort to the query's informational density. The intensified local response in fine-grained experts underscores the model's ability to capture high-frequency visual signals when precise matching is required, while the weaker, broader response in coarse-grained experts reflects its resilience to the inherent ambiguity of sparse descriptions.

\begin{figure}[t]
\centering
\includegraphics[width=0.98\linewidth]{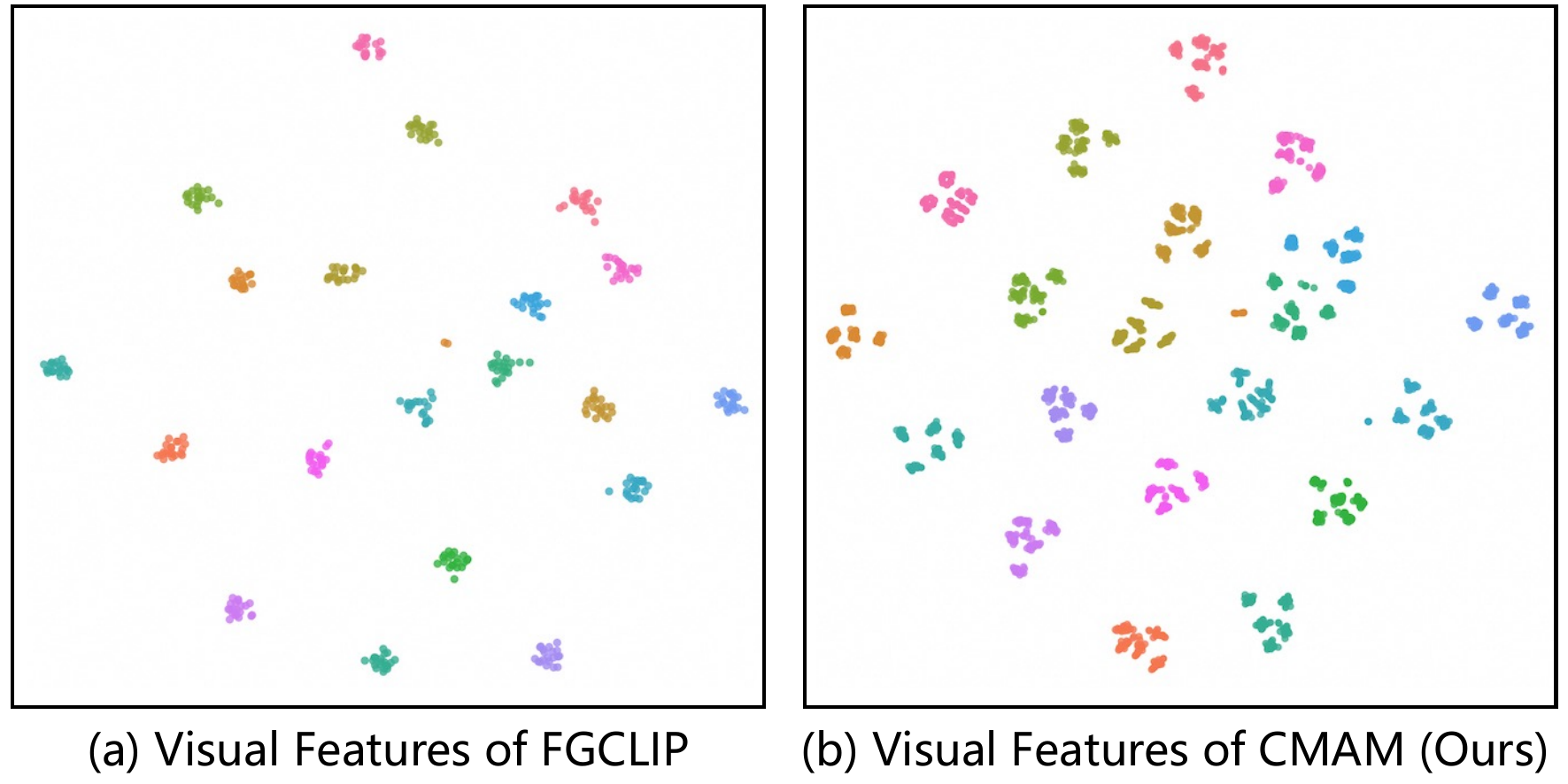}
\caption{T-SNE visualization of visual features for 20 identities. (a) FGCLIP features cluster solely by identity. (b) CMAM features exhibit a hierarchical structure where each identity cluster is further stratified into five sub-clusters, demonstrating effective intra-identity granularity disentanglement. }
\label{fig:dist}
\end{figure}

\noindent
\textbf{Analysis of Visual Feature Distribution.} Fig.~\ref{fig:dist} compares the feature manifolds of FGCLIP~\cite{xie2025fgclip} and CMAM via t-SNE for 20 identities. In FGCLIP, features aggregate strictly by identity in a uni-granular fashion. Conversely, CMAM displays a clear hierarchical topology: while identities remain well-separated, each identity cluster further partitions into five distinct sub-clusters corresponding to the granularity levels. This nested structure confirms that CMAM achieves effective intra-identity granularity disentanglement, preserving robust discriminative cues while explicitly encoding the specificity of visual signals. Such a distribution validates the framework's ability to navigate the complex trade-off between identity consistency and granularity-aware retrieval.

\begin{figure}[t]
\centering
\includegraphics[width=\linewidth]{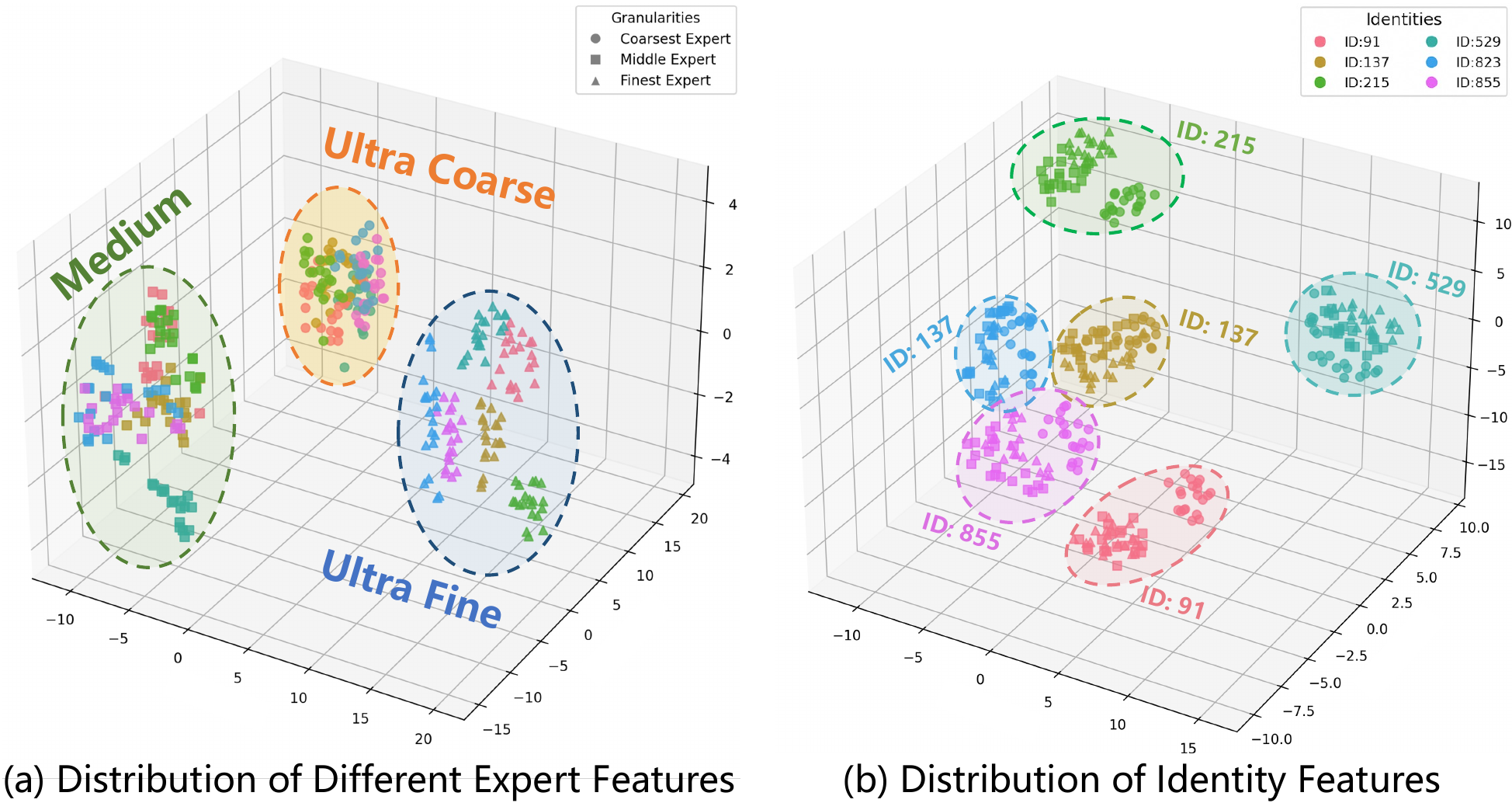}
\caption{3D t-SNE visualization of expert and global manifolds for six identities. (a) Expert-specific feature vectors demonstrate distinct clustering and functional disentanglement. (b) Global representations show clear identity separation while capturing the semantic proximity of visually similar individuals.}
\vspace{-3mm}
\label{fig:dist_expert}
\end{figure}

\noindent
\textbf{Manifold Analysis of Expert and Identity Features.} To explore expert specialization and identity discriminability, we visualize the feature manifolds of six randomly selected identities using 3D t-SNE (Fig.~\ref{fig:dist_expert}).
First, we analyze the expert-specific feature vectors. As shown in (a), features from different experts exhibit distinct spatial segregation across the manifold, confirming clear functional disentanglement: each expert operates in a specialized subspace to capture granularity-specific signals.
Second, we project the global multi-grained representations in (b). While different identities are overall well-partitioned, we observe semantic proximity among visually similar individuals. This dual-structure manifold validates that CMAM not only achieves precise expert specialization but also constructs a robust identity-aware semantic space that effectively reconciles identity-level discriminability with cross-identity similarity.

\section{Conclusion}
This paper formalizes the paradigm of text-based person retrieval with any granularity, bridging the gap between constrained benchmarks and unpredictable real-world queries. We introduced UFine6926-MG, the first granularity-controlled dataset, and MG-Eval, a multi-grained benchmark incorporating cross-identity labels to better reflect the inherent ambiguity of varied descriptions. To address these challenges, we proposed the CMAM framework, which integrates orthogonal-expert perception, probabilistic alignment, and granularity-consistent reasoning to learn robust, multi-grained representations. Extensive experiments demonstrate that CMAM achieves state-of-the-art performance and superior adaptability across the granularity spectrum. As a systematic extension of our prior work, this research establishes a foundational framework for developing more flexible and reliable person retrieval systems in complex, large-scale applications.

\bibliographystyle{IEEEtran}
\bibliography{ref}

@String { ACMMM        = {ACM International Conference on Multimedia} }

@String { ARXIV        = {arXiv} }

@String { BMVC         = {British Machine Vision Conference} }

@String { CVPR         = {IEEE/CVF Conference on Computer Vision and Pattern Recognition} }

@String { ECCV         = {European Conference on Computer Vision} }

@String { AAAI         = {AAAI Conference on Artificial Intelligence} }

@String { ICASSP       = {IEEE International Conference on Acoustics, Speech, and Signal Processing} }

@String { ICCV         = {IEEE International Conference on Computer Vision} }

@String { ICML         = {International Conference on Machine Learning} }

@String { NIPS         = {Neural Information Processing Systems} }

@String { TIP          = {IEEE Transactions on Image Processing} }

@String { WACV         = {IEEE Winter Conference on Applications of Computer Vision}}

@String { ACMMM        = {ACMMM} }

@String { IJCAI        = {IJCAI} }

@String { BMVC         = {BMVC} }

@String { CVPR         = {CVPR} }

@String { ECCV         = {ECCV} }

@String { AAAI         = {AAAI} }

@String { ICASSP       = {ICASSP} }

@String { ICCV         = {ICCV} }

@String { ICML         = {ICML} }

@String { NIPS         = {NeurIPS} }

@String { WACV         = {WACV}}

@inproceedings{cuhkpedes,
  title={Person search with natural language description},
  author={Li, Shuang and Xiao, Tong and Li, Hongsheng and Zhou, Bolei and Yue, Dayu and Wang, Xiaogang},
  booktitle=CVPR,
  pages={1970--1979},
  year={2017}
}

@article{icfgpedes,
  title={Semantically self-aligned network for text-to-image part-aware person re-identification},
  author={Ding, Zefeng and Ding, Changxing and Shao, Zhiyin and Tao, Dacheng},
  journal={arXiv preprint arXiv:2107.12666},
  year={2021}
}

@inproceedings{rstpreid,
  title={Dssl: Deep surroundings-person separation learning for text-based person retrieval},
  author={Zhu, Aichun and Wang, Zijie and Li, Yifeng and Wan, Xili and Jin, Jing and Wang, Tian and Hu, Fangqiang and Hua, Gang},
  booktitle=ACMMM,
  pages={209--217},
  year={2021}
}

@inproceedings{ufinebench,
  title={Ufinebench: Towards text-based person retrieval with ultra-fine granularity},
  author={Zuo, Jialong and Zhou, Hanyu and Nie, Ying and Zhang, Feng and Guo, Tianyu and Sang, Nong and Wang, Yunhe and Gao, Changxin},
  booktitle=CVPR,
  pages={22010--22019},
  year={2024}
}

@article{zuo2024plip,
  title={Plip: Language-image pre-training for person representation learning},
  author={Zuo, Jialong and Hong, Jiahao and Zhang, Feng and Yu, Changqian and Zhou, Hanyu and Gao, Changxin and Sang, Nong and Wang, Jingdong},
  journal=NIPS,
  volume={37},
  pages={45666--45702},
  year={2024}
}

@inproceedings{mals,
  title={Towards unified text-based person retrieval: A large-scale multi-attribute and language search benchmark},
  author={Yang, Shuyu and Zhou, Yinan and Zheng, Zhedong and Wang, Yaxiong and Zhu, Li and Wu, Yujiao},
  booktitle=ACMMM,
  pages={4492--4501},
  year={2023}
}

@inproceedings{zhang2018deep,
  title={Deep cross-modal projection learning for image-text matching},
  author={Zhang, Ying and Lu, Huchuan},
  booktitle=ECCV,
  pages={686--701},
  year={2018}
}

@article{zheng2020dual,
  title={Dual-path convolutional image-text embeddings with instance loss},
  author={Zheng, Zhedong and Zheng, Liang and Garrett, Michael and Yang, Yi and Xu, Mingliang and Shen, Yi-Dong},
  journal={ACM Transactions on Multimedia Computing, Communications, and Applications (TOMM)},
  volume={16},
  number={2},
  pages={1--23},
  year={2020},
  publisher={ACM New York, NY, USA}
}

@inproceedings{chen2018improving,
  title={Improving text-based person search by spatial matching and adaptive threshold},
  author={Chen, Tianlang and Xu, Chenliang and Luo, Jiebo},
  booktitle=WACV,
  pages={1879--1887},
  year={2018},
  organization={IEEE}
}

@inproceedings{wang2019language,
  title={Language person search with mutually connected classification loss},
  author={Wang, Yuyu and Bo, Chunjuan and Wang, Dong and Wang, Shuang and Qi, Yunwei and Lu, Huchuan},
  booktitle=ICASSP,
  pages={2057--2061},
  year={2019},
  organization={IEEE}
}

@inproceedings{jing2020pose,
  title={Pose-guided multi-granularity attention network for text-based person search},
  author={Jing, Ya and Si, Chenyang and Wang, Junbo and Wang, Wei and Wang, Liang and Tan, Tieniu},
  booktitle=AAAI,
  volume={34},
  number={07},
  pages={11189--11196},
  year={2020}
}

@inproceedings{wang2020vitaa,
  title={Vitaa: Visual-textual attributes alignment in person search by natural language},
  author={Wang, Zhe and Fang, Zhiyuan and Wang, Jun and Yang, Yezhou},
  booktitle=ECCV,
  pages={402--420},
  year={2020},
  organization={Springer}
}

@article{niu2020improving,
  title={Improving description-based person re-identification by multi-granularity image-text alignments},
  author={Niu, Kai and Huang, Yan and Ouyang, Wanli and Wang, Liang},
  journal=TIP,
  volume={29},
  pages={5542--5556},
  year={2020},
  publisher={IEEE}
}

@inproceedings{wang2022caibc,
  title={Caibc: Capturing all-round information beyond color for text-based person retrieval},
  author={Wang, Zijie and Zhu, Aichun and Xue, Jingyi and Wan, Xili and Liu, Chao and Wang, Tian and Li, Yifeng},
  booktitle=ACMMM,
  pages={5314--5322},
  year={2022}
}

@inproceedings{wu2021lapscore,
  title={Lapscore: language-guided person search via color reasoning},
  author={Wu, Yushuang and Yan, Zizheng and Han, Xiaoguang and Li, Guanbin and Zou, Changqing and Cui, Shuguang},
  booktitle=ICCV,
  pages={1624--1633},
  year={2021}
}

@inproceedings{irra,
  title={Cross-modal implicit relation reasoning and aligning for text-to-image person retrieval},
  author={Jiang, Ding and Ye, Mang},
  booktitle=CVPR,
  pages={2787--2797},
  year={2023}
}

@inproceedings{rde,
  title={Noisy-correspondence learning for text-to-image person re-identification},
  author={Qin, Yang and Chen, Yingke and Peng, Dezhong and Peng, Xi and Zhou, Joey Tianyi and Hu, Peng},
  booktitle=CVPR,
  pages={27197--27206},
  year={2024}
}

@inproceedings{cao2024empirical,
  title={An empirical study of clip for text-based person search},
  author={Cao, Min and Bai, Yang and Zeng, Ziyin and Ye, Mang and Zhang, Min},
  booktitle=AAAI,
  volume={38},
  number={1},
  pages={465--473},
  year={2024}
}

@article{yan2023clip,
  title={Clip-driven fine-grained text-image person re-identification},
  author={Yan, Shuanglin and Dong, Neng and Zhang, Liyan and Tang, Jinhui},
  journal=TIP,
  volume={32},
  pages={6032--6046},
  year={2023},
  publisher={IEEE}
}

@article{wang2026p,
  title={P-CLIP: Progressive Discrepancy Learning for One-Shot Text-to-Image Person Re-identification},
  author={Wang, Chengji and Dong, Ming and Ye, Mang and Sun, Hao and Jiang, Xingpeng},
  journal=TIP,
  year={2026},
  publisher={IEEE}
}

@inproceedings{clip,
  title={Learning transferable visual models from natural language supervision},
  author={Radford, Alec and Kim, Jong Wook and Hallacy, Chris and Ramesh, Aditya and Goh, Gabriel and Agarwal, Sandhini and Sastry, Girish and Askell, Amanda and Mishkin, Pamela and Clark, Jack and others},
  booktitle=ICML,
  pages={8748--8763},
  year={2021},
  organization={PmLR}
}

@inproceedings{liu2024clip,
  title={Clip-based synergistic knowledge transfer for text-based person retrieval},
  author={Liu, Yating and Li, Yaowei and Liu, Zimo and Yang, Wenming and Wang, Yaowei and Liao, Qingmin},
  booktitle=ICASSP,
  pages={7935--7939},
  year={2024},
  organization={IEEE}
}

@article{shen2025enhancing,
  title={Enhancing visual representation for text-based person searching},
  author={Shen, Wei and Fang, Ming and Wang, Yuxia and Xiao, Jiafeng and Li, Diping and Chen, Huangqun and Xu, Ling and Zhang, Weifeng},
  journal={Knowledge-Based Systems},
  volume={309},
  pages={112893},
  year={2025},
  publisher={Elsevier}
}

@inproceedings{bai2025chat,
  title={Chat-based Person Retrieval via Dialogue-Refined Cross-Modal Alignment},
  author={Bai, Yang and Ji, Yucheng and Cao, Min and Wang, Jinqiao and Ye, Mang},
  booktitle=CVPR,
  pages={3952--3962},
  year={2025}
}

@inproceedings{qin2025human,
  title={Human-centered Interactive Learning via MLLMs for Text-to-Image Person Re-identification},
  author={Qin, Yang and Chen, Chao and Fu, Zhihang and Peng, Dezhong and Peng, Xi and Hu, Peng},
  booktitle=CVPR,
  pages={14390--14399},
  year={2025}
}

@inproceedings{
lu2025llavareid,
title={{LL}a{VA}-Re{ID}: Selective Multi-image Questioner for Interactive Person Re-Identification},
author={Yiding Lu and Mouxing Yang and Dezhong Peng and Peng Hu and Yijie Lin and Xi Peng},
booktitle=ICML,
year={2025},
}

@inproceedings{
xie2025fgclip,
title={{FG}-{CLIP}: Fine-Grained Visual and Textual Alignment},
author={Chunyu Xie and Bin Wang and Fanjing Kong and Jincheng Li and Dawei Liang and Gengshen Zhang and Dawei Leng and Yuhui Yin},
booktitle=ICML,
year={2025}
}

@inproceedings{fujii2023bilma,
  title={Bilma: Bidirectional local-matching for text-based person re-identification},
  author={Fujii, Takuro and Tarashima, Shuhei},
  booktitle=ICCV,
  pages={2786--2790},
  year={2023}
}

@inproceedings{shao2022learning,
  title={Learning granularity-unified representations for text-to-image person re-identification},
  author={Shao, Zhiyin and Zhang, Xinyu and Fang, Meng and Lin, Zhifeng and Wang, Jian and Ding, Changxing},
  booktitle=ACMMM,
  pages={5566--5574},
  year={2022}
}

@inproceedings{bai2023rasa,
  title={RaSa: relation and sensitivity aware representation learning for text-based person search},
  author={Bai, Yang and Cao, Min and Gao, Daming and Cao, Ziqiang and Chen, Chen and Fan, Zhenfeng and Nie, Liqiang and Zhang, Min},
  booktitle=IJCAI,
  pages={555--563},
  year={2023}
}

@article{gou2025instance,
  title={Instance-level feature bias calibration learning for text-to-image person re-identification},
  author={Gou, Yifeng and Li, Ziqiang and Zhang, Junyin and Wang, Yunnan and Ge, Yongxin},
  journal={Knowledge-Based Systems},
  volume={315},
  pages={113251},
  year={2025},
  publisher={Elsevier}
}

@article{zeng2025hierarchical,
  title={Hierarchical knowledge-guided reasoning for text-based person re-identification},
  author={Zeng, Ruigeng and Ma, Wentao and Zhou, Tongqing and Zhao, Shan and Mao, Xinjun and Liu, Jie},
  journal={Neural Networks},
  pages={107888},
  year={2025},
  publisher={Elsevier}
}

@inproceedings{zheng2015scalable,
  title={Scalable person re-identification: A benchmark},
  author={Zheng, Liang and Shen, Liyue and Tian, Lu and Wang, Shengjin and Wang, Jingdong and Tian, Qi},
  booktitle=ICCV,
  pages={1116--1124},
  year={2015}
}

@inproceedings{wei2018person,
  title={Person transfer gan to bridge domain gap for person re-identification},
  author={Wei, Longhui and Zhang, Shiliang and Gao, Wen and Tian, Qi},
  booktitle=CVPR,
  pages={79--88},
  year={2018}
}

@inproceedings{he2021transreid,
  title={Transreid: Transformer-based object re-identification},
  author={He, Shuting and Luo, Hao and Wang, Pichao and Wang, Fan and Li, Hao and Jiang, Wei},
  booktitle=ICCV,
  pages={15013--15022},
  year={2021}
}

@article{zuo2024cross,
  title={Cross-video identity correlating for person re-identification pre-training},
  author={Zuo, Jialong and Nie, Ying and Zhou, Hanyu and Zhang, Huaxin and Wang, Haoyu and Guo, Tianyu and Sang, Nong and Gao, Changxin},
  journal=NIPS,
  volume={37},
  pages={25228--25250},
  year={2024}
}

@article{ye2021deep,
  title={Deep learning for person re-identification: A survey and outlook},
  author={Ye, Mang and Shen, Jianbing and Lin, Gaojie and Xiang, Tao and Shao, Ling and Hoi, Steven CH},
  journal={IEEE transactions on pattern analysis and machine intelligence},
  volume={44},
  number={6},
  pages={2872--2893},
  year={2021},
  publisher={IEEE}
}

@article{ye2025transformer,
  title={Transformer for object re-identification: A survey},
  author={Ye, Mang and Chen, Shuoyi and Li, Chenyue and Zheng, Wei-Shi and Crandall, David and Du, Bo},
  journal={International Journal of Computer Vision},
  volume={133},
  number={5},
  pages={2410--2440},
  year={2025},
  publisher={Springer}
}

@article{yang2025qwen3,
  title={Qwen3 technical report},
  author={Yang, An and Li, Anfeng and Yang, Baosong and Zhang, Beichen and Hui, Binyuan and Zheng, Bo and Yu, Bowen and Gao, Chang and Huang, Chengen and Lv, Chenxu and others},
  journal={arXiv preprint arXiv:2505.09388},
  year={2025}
}

@article{bai2025qwen2,
  title={Qwen2.5-vl technical report},
  author={Bai, Shuai and Chen, Keqin and Liu, Xuejing and Wang, Jialin and Ge, Wenbin and Song, Sibo and Dang, Kai and Wang, Peng and Wang, Shijie and Tang, Jun and others},
  journal={arXiv preprint arXiv:2502.13923},
  year={2025}
}

@article{faghri2018vse++,
  title={VSE++: Improving Visual-Semantic Embeddings with Hard Negatives},
  author={Faghri, Fartash and Fleet, David J and Kiros, Jamie Ryan and Fidler, Sanja},
  booktitle = {BMVC},
  year={2018}
}

@inproceedings{chen2020imram,
  title={Imram: Iterative matching with recurrent attention memory for cross-modal image-text retrieval},
  author={Chen, Hui and Ding, Guiguang and Liu, Xudong and Lin, Zijia and Liu, Ji and Han, Jungong},
  booktitle=CVPR,
  pages={12655--12663},
  year={2020}
}

@inproceedings{pan2023fine,
  title={Fine-grained image-text matching by cross-modal hard aligning network},
  author={Pan, Zhengxin and Wu, Fangyu and Zhang, Bailing},
  booktitle=CVPR,
  pages={19275--19284},
  year={2023}
}

@inproceedings{xiao2025flair,
  title={Flair: Vlm with fine-grained language-informed image representations},
  author={Xiao, Rui and Kim, Sanghwan and Georgescu, Mariana-Iuliana and Akata, Zeynep and Alaniz, Stephan},
  booktitle=CVPR,
  pages={24884--24894},
  year={2025}
}

@article{xiao2025pixclip,
  title={PixCLIP: Achieving Fine-grained Visual Language Understanding via Any-granularity Pixel-Text Alignment Learning},
  author={Xiao, Yicheng and Chen, Yu and Ma, Haoxuan and Hong, Jiale and Li, Caorui and Wu, Lingxiang and Guo, Haiyun and Wang, Jinqiao},
  journal={arXiv preprint arXiv:2511.04601},
  year={2025}
}

@article{truong2025mulclip,
  title={MulCLIP: A Multi-level Alignment Framework for Enhancing Fine-grained Long-context CLIP},
  author={Truong, Chau and Quang, Hieu Ta and Le, Dung D},
  journal={arXiv preprint arXiv:2512.07128},
  year={2025}
}

@inproceedings{ma2022x,
  title={X-clip: End-to-end multi-grained contrastive learning for video-text retrieval},
  author={Ma, Yiwei and Xu, Guohai and Sun, Xiaoshuai and Yan, Ming and Zhang, Ji and Ji, Rongrong},
  booktitle=ACMMM,
  pages={638--647},
  year={2022}
}

@inproceedings{huang2025mgsgm,
  title={MGSGM: Multi-Granularity Selective Graph Mamba for Image-Text Retrieval},
  author={Huang, Yongle and Bu, Yongfeng and Guo, Keyu and Liu, Zedong and Song, Xiangyu and Sun, Shijie},
  booktitle={Proceedings of the 2025 International Conference on Multimedia Retrieval},
  pages={1983--1987},
  year={2025}
}

@inproceedings{zeng2022multi,
  title={Multi-Grained Vision Language Pre-Training: Aligning Texts with Visual Concepts},
  author={Zeng, Yan and Zhang, Xinsong and Li, Hang},
  booktitle=ICML,
  pages={25994--26009},
  year={2022},
  organization={PMLR}
}

@inproceedings{hendriksen2025benchmark,
  title={Benchmark Granularity and Model Robustness for Image-Text Retrieval: A Reproducibility Study},
  author={Hendriksen, Mariya and Zhang, Shuo and Reinanda, Ridho and Yahya, Mohamed and Meij, Edgar and de Rijke, Maarten},
  booktitle={Proceedings of the 48th International ACM SIGIR Conference on Research and Development in Information Retrieval},
  pages={3183--3193},
  year={2025}
}

@article{li2021align,
  title={Align before fuse: Vision and language representation learning with momentum distillation},
  author={Li, Junnan and Selvaraju, Ramprasaath and Gotmare, Akhilesh and Joty, Shafiq and Xiong, Caiming and Hoi, Steven Chu Hong},
  journal=NIPS,
  volume={34},
  pages={9694--9705},
  year={2021}
}

@inproceedings{li2022blip,
  title={Blip: Bootstrapping language-image pre-training for unified vision-language understanding and generation},
  author={Li, Junnan and Li, Dongxu and Xiong, Caiming and Hoi, Steven},
  booktitle=ICML,
  pages={12888--12900},
  year={2022},
  organization={PMLR}
}

@inproceedings{
yao2022filip,
title={{FILIP}: Fine-grained Interactive Language-Image Pre-Training},
author={Lewei Yao and Runhui Huang and Lu Hou and Guansong Lu and Minzhe Niu and Hang Xu and Xiaodan Liang and Zhenguo Li and Xin Jiang and Chunjing Xu},
booktitle=ICML,
year={2022}
}

@inproceedings{zhang2024long,
  title={Long-clip: Unlocking the long-text capability of clip},
  author={Zhang, Beichen and Zhang, Pan and Dong, Xiaoyi and Zang, Yuhang and Wang, Jiaqi},
  booktitle=ECCV,
  pages={310--325},
  year={2024},
  organization={Springer}
}

@article{zhang2021fairmot,
  title={Fairmot: On the fairness of detection and re-identification in multiple object tracking},
  author={Zhang, Yifu and Wang, Chunyu and Wang, Xinggang and Zeng, Wenjun and Liu, Wenyu},
  journal={International journal of computer vision},
  volume={129},
  number={11},
  pages={3069--3087},
  year={2021},
  publisher={Springer}
}

@inproceedings{wu2017rgb,
  title={RGB-infrared cross-modality person re-identification},
  author={Wu, Ancong and Zheng, Wei-Shi and Yu, Hong-Xing and Gong, Shaogang and Lai, Jianhuang},
  booktitle=ICCV,
  pages={5380--5389},
  year={2017}
}

@article{yu2026i2id,
  title={I2ID: Disentangling identity features via synchronized masking for zero-shot composed person retrieval},
  author={Yu, Guo and Wang, Di and Yan, Chengwei and Yan, Feng and Luo, Nan and Wang, Yifeng and Wang, Quan},
  journal={Pattern Recognition},
  pages={113654},
  year={2026},
  publisher={Elsevier}
}

@inproceedings{wang2024fine,
author = {Wang, Di and Yan, Feng and Wang, Yifeng and Zhao, Lin and Liang, Xiao and Zhong, Haodi and Zhang, Ronghua},
title = {Fine-grained Semantics-aware Representation Learning for Text-based Person Retrieval},
year = {2024},
booktitle = {ICMR},
pages = {92–100},
numpages = {9},
}

@inproceedings{zhang2023diverse,
  title={Diverse embedding expansion network and low-light cross-modality benchmark for visible-infrared person re-identification},
  author={Zhang, Yukang and Wang, Hanzi},
  booktitle=CVPR,
  pages={2153--2162},
  year={2023}
}

@article{vaswani2017attention,
  title={Attention is all you need},
  author={Vaswani, Ashish and Shazeer, Noam and Parmar, Niki and Uszkoreit, Jakob and Jones, Llion and Gomez, Aidan N and Kaiser, {\L}ukasz and Polosukhin, Illia},
  journal=NIPS,
  volume={30},
  year={2017}
}

@inproceedings{tan2024harnessing,
  title={Harnessing the power of mllms for transferable text-to-image person reid},
  author={Tan, Wentan and Ding, Changxing and Jiang, Jiayu and Wang, Fei and Zhan, Yibing and Tao, Dapeng},
  booktitle=CVPR,
  pages={17127--17137},
  year={2024}
}

@inproceedings{jiang2025modeling,
  title={Modeling Thousands of Human Annotators for Generalizable Text-to-Image Person Re-identification},
  author={Jiang, Jiayu and Ding, Changxing and Tan, Wentao and Wang, Junhong and Tao, Jin and Xu, Xiangmin},
  booktitle=CVPR,
  pages={9220--9230},
  year={2025}
}

\begin{IEEEbiography}[{\includegraphics[width=1in,height=1.25in,clip,keepaspectratio]{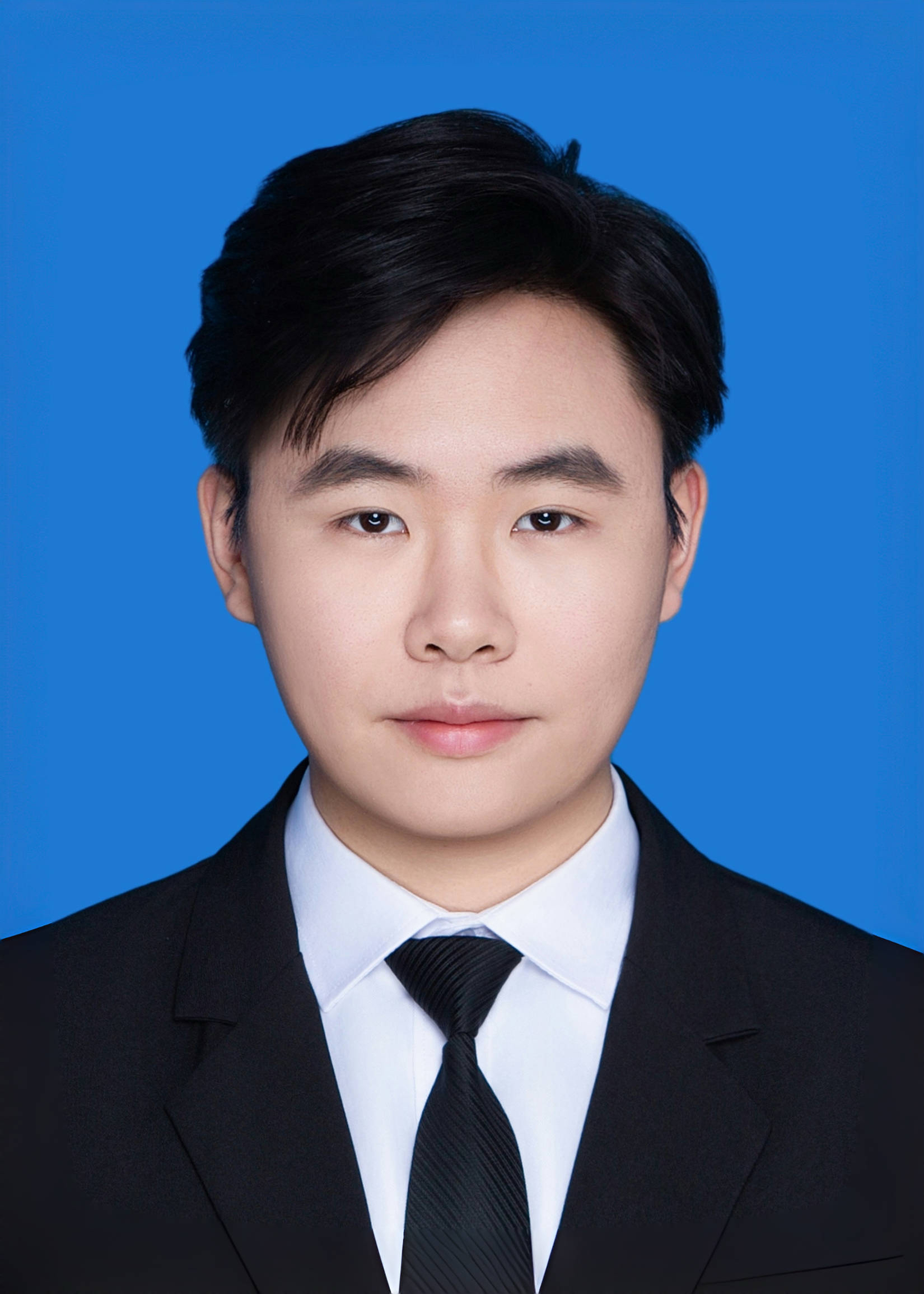}}]{Jialong Zuo}
received B.S. degree in Huazhong University of Science and Technology, China, in 2022. He is currently pursuing the Ph.D. degree in the School of Artificial Intelligence and Automation, Huazhong University of Science and Technology, supervised by Prof. Changxin Gao. His research interests include person re-identification, representation learning, and multimodal learning.
\end{IEEEbiography}

\begin{IEEEbiography}[{\includegraphics[width=1in,height=1.25in,clip,keepaspectratio]{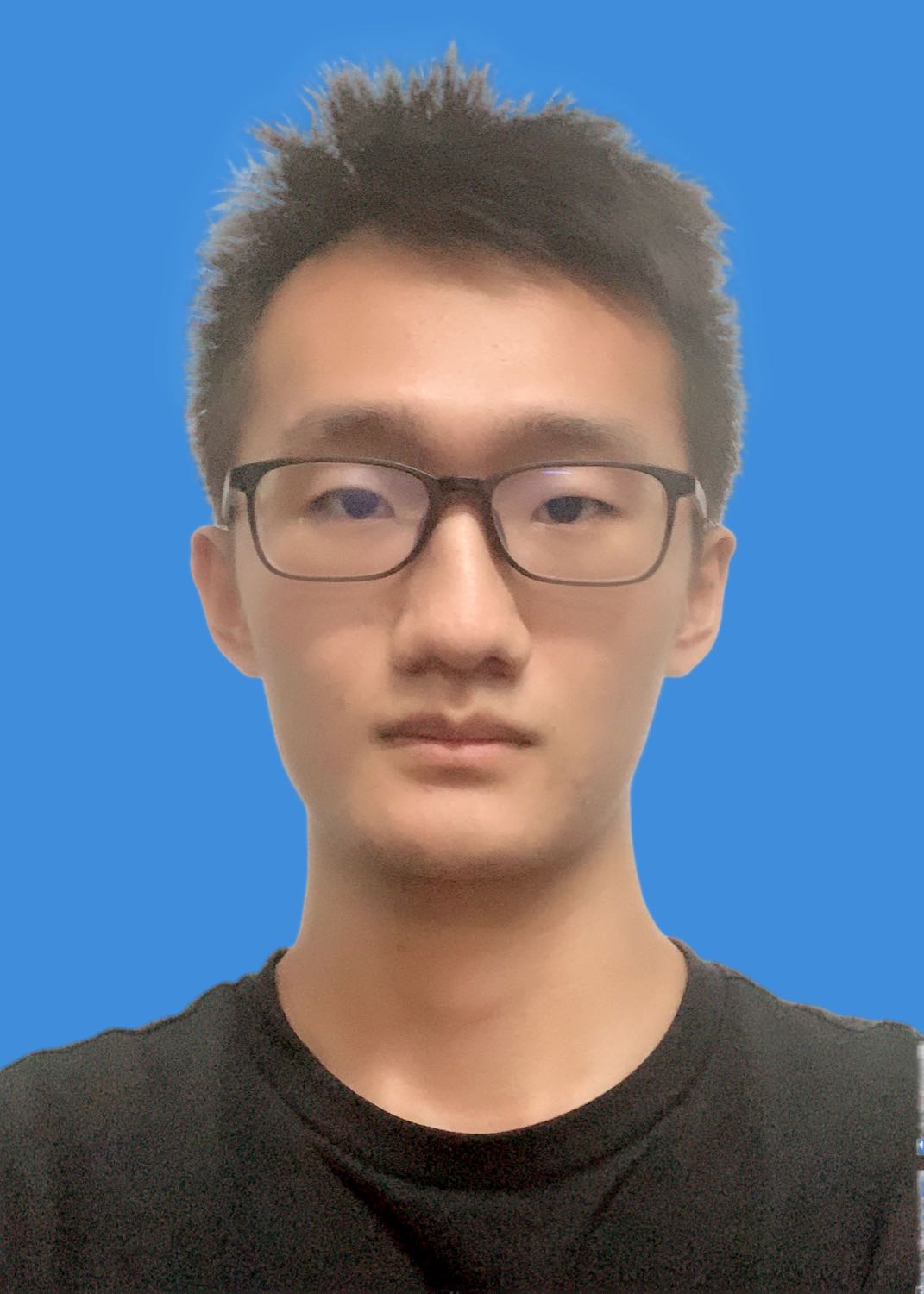}}]{Hanyu Zhou}
received B.S. degree in Huazhong University of Science and Technology, China, in 2023. He is currently pursuing the Ph.D. degree in the School of Artificial Intelligence and Automation, Huazhong University of Science and Technology, supervised by Prof. Changxin Gao. His research interest is person re-identification and text-based person retrieval.
\end{IEEEbiography}

\begin{IEEEbiography}[{\includegraphics[width=1in,height=1.25in,clip,keepaspectratio]{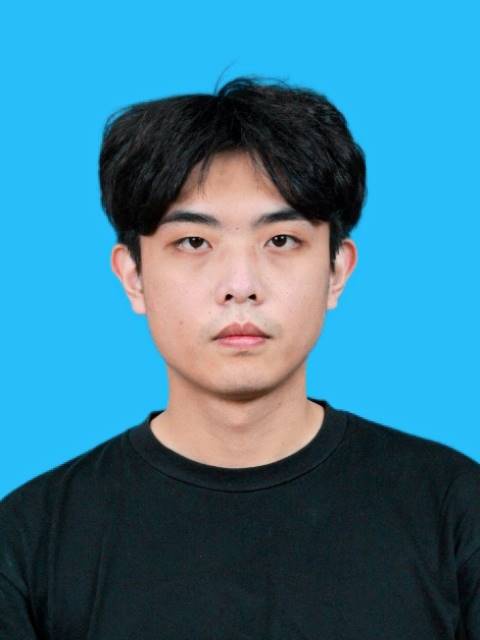}}]{Dongyue Wu}
received B.S. degree in Huazhong University of Science and Technology, China, in 2020. He is currently pursuing the Ph.D. degree in the School of Artificial Intelligence and Automation, Huazhong University of Science and Technology, supervised by Prof. Changxin Gao. His research interests include dataset pruning, network pruning, training acceleration and multimodal learning.
\end{IEEEbiography}

\begin{IEEEbiography}[{\includegraphics[width=1in,height=1.25in,clip,keepaspectratio]{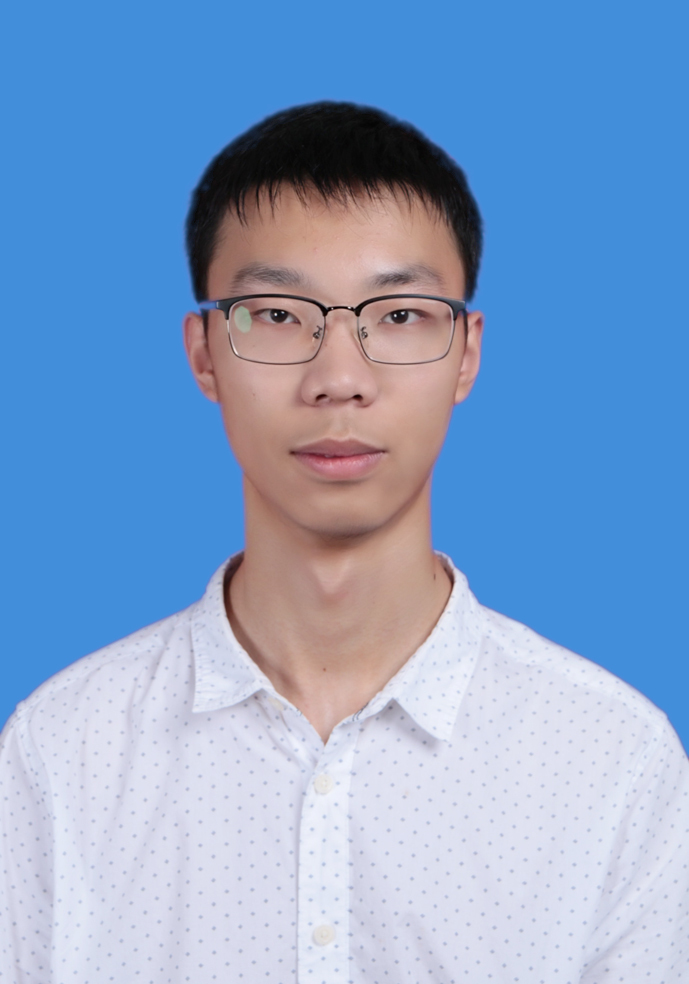}}]{Yongtai Deng}
received B.S. degree in Huazhong University of Science and Technology, China, in 2025. He is currently pursuing the Master degree in the School of Artificial Intelligence and Automation, Huazhong University of Science and Technology, supervised by Prof. Changxin Gao. His research interest is person re-identification and text-based person retrieval.
\end{IEEEbiography}

\begin{IEEEbiography}[{\includegraphics[width=1in,height=1.25in,clip,keepaspectratio]{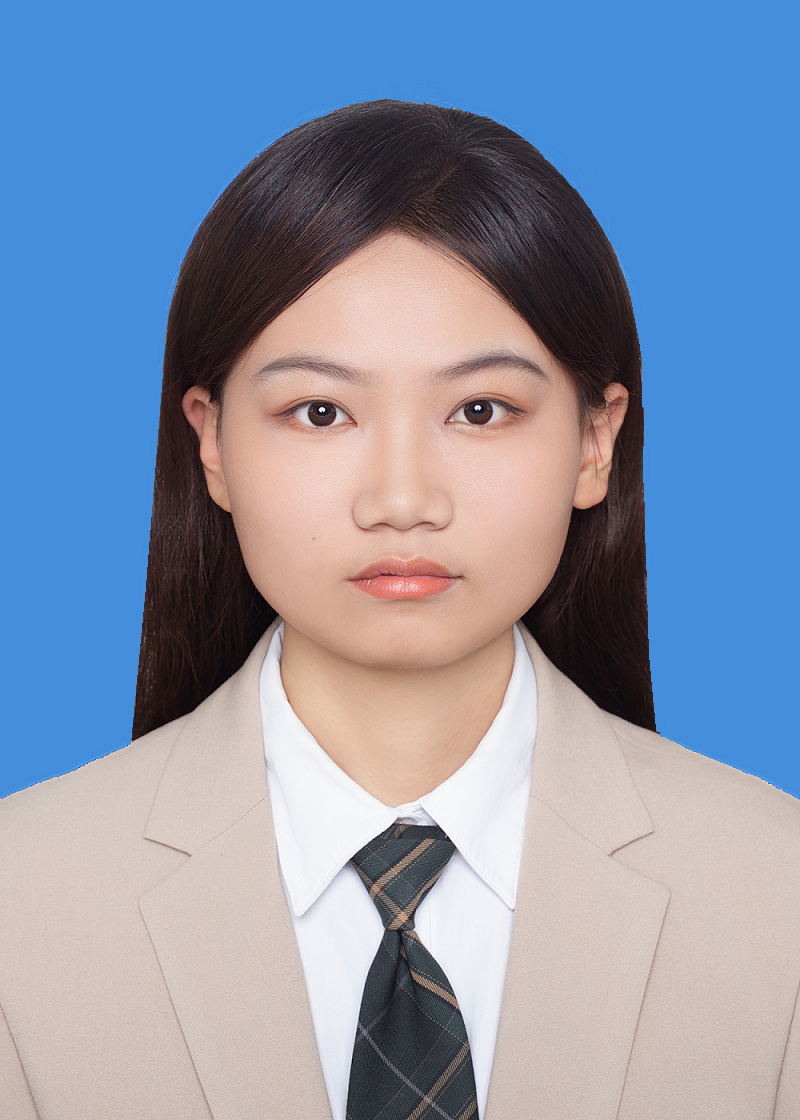}}]{Mengdan Tan}
received B.S. degree in Huazhong University of Science and Technology, China, in 2024. She is currently pursuing the Master degree in the School of Artificial Intelligence and Automation, Huazhong University of Science and Technology, supervised by Prof. Changxin Gao. Her research interest is person re-identification and text-based person retrieval.
\end{IEEEbiography}

\begin{IEEEbiography}[{\includegraphics[width=1in,height=1.25in,clip,keepaspectratio]{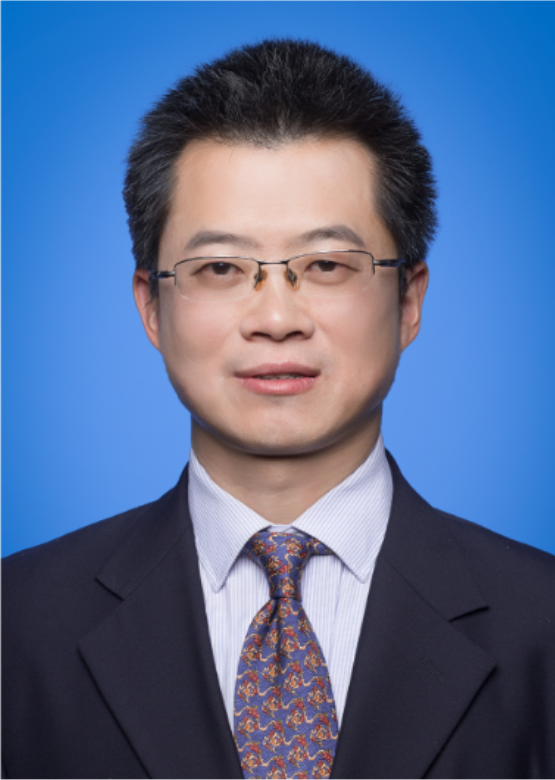}}]{Nong Sang}
(Member, IEEE) received the Ph.D. degree in pattern recognition and intelligent systems from Huazhong University of Science and Technology in 2000. He is currently a professor at the School of Artificial Intelligence and Automation, Huazhong University of Science and Technology, China. His research interests include object detection and recognition, object tracking, image/video semantic segmentation, intelligent processing and analysis of surveillance videos.
\end{IEEEbiography}

\begin{IEEEbiography}[{\includegraphics[width=1in,height=1.25in,clip,keepaspectratio]{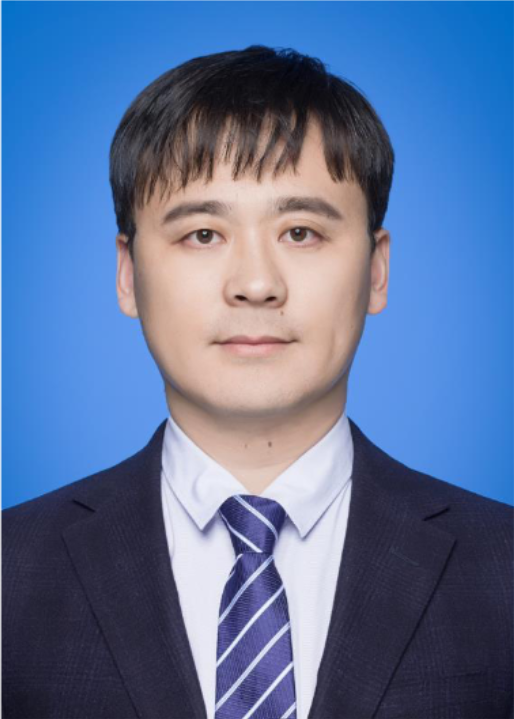}}]{Changxin Gao}
(Senior Member, IEEE) received the B.S. degree in optical information science and technology and the Ph.D. degree in pattern recognition and intelligent systems, both from Huazhong University of Science and Technology in 2005 and 2010. He is currently a professor at the School of Artificial Intelligence and Automation, Huazhong University of Science and Technology, China. From 2012 to 2013, he was a research assistant in Universit\'{e} Catholique de Louvain, Belgium. His research interests are image understanding and surveillance video analysis.
\end{IEEEbiography}

\begin{IEEEbiography}[{\includegraphics[width=1in,height=1.25in,clip,keepaspectratio]{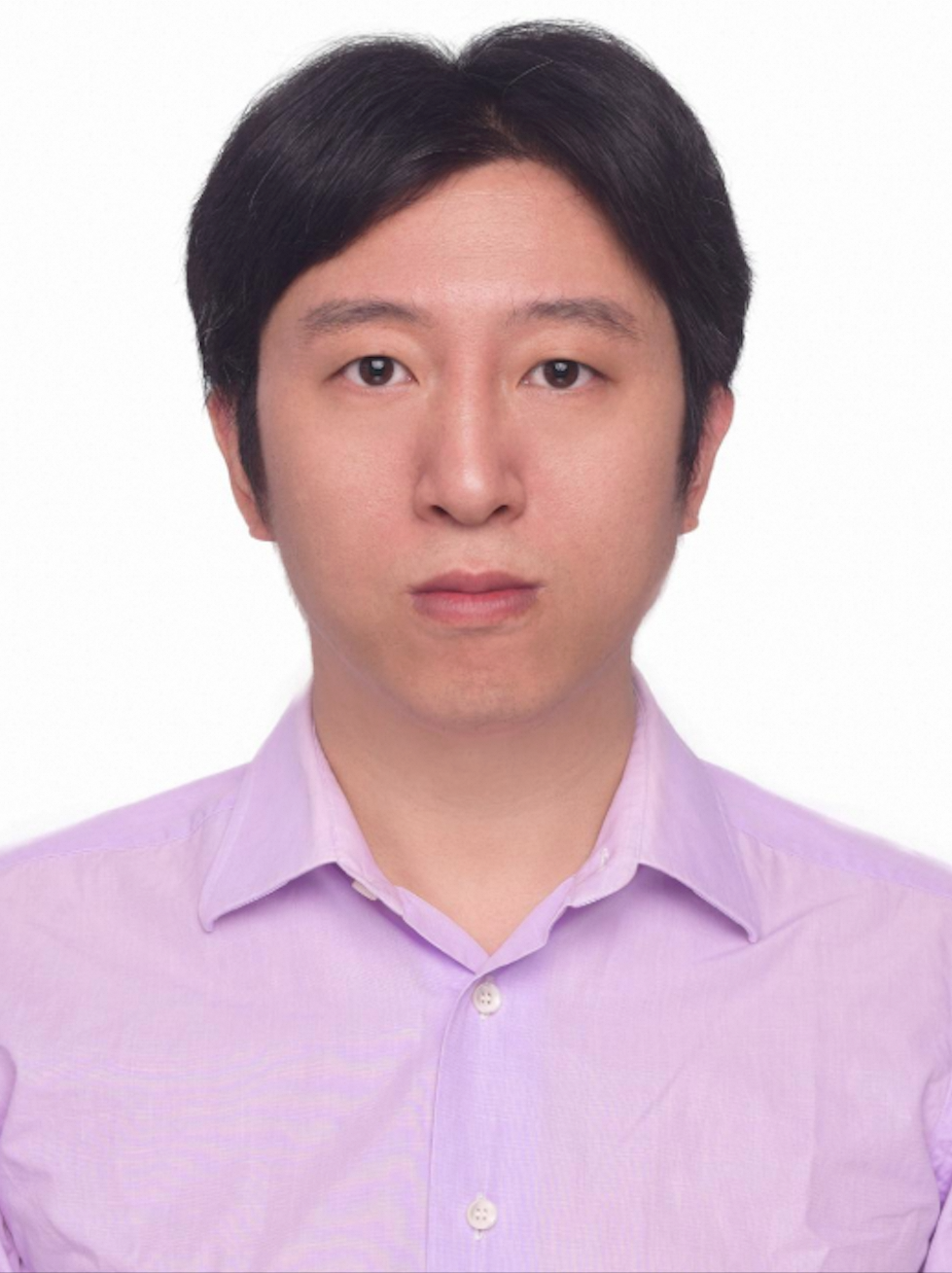}}]{Xiang Bai}
(Fellow, IEEE) received the BS, MS, and PhD degrees in electronics and information engineering from the Huazhong University of Science and Technology (HUST), Wuhan, China, in 2003, 2005, and 2009, respectively. He is currently a professor with the School of Software, HUST. He is also the vice-director of the National Center of AntiCounterfeiting Technology, HUST. His research interests include object recognition, shape analysis, scene text recognition, and intelligent systems. He is also an associate editor for IEEE TRANSACTIONS ON PATTERN ANALYSIS AND MACHINE INTELLIGENCE, and an associate editor-in-chief for Pattern Recognition. He is a fellow of IEEE and IAPR.
\end{IEEEbiography}

\end{document}